\documentclass[twoside,11pt]{article}

\usepackage{jmlr2e}
\usepackage{microtype}
\usepackage{mathtools}

\usepackage{tikz}
\usepackage{pgfplots}
\usepackage{amsmath}
\usepackage{amssymb}
\usepackage{graphicx}
\usepackage{MnSymbol}
\usepackage{pifont}
\usepackage{booktabs}

\newif\iffinal 

\ShortHeadings{Dual Control for Approximate Bayesian Reinforcement
Learning}{Klenske and Hennig}
\firstpageno{1}

\definecolor{lred}{RGB}{200,0,0}
\definecolor{dred}{RGB}{130,0,0}
\definecolor{dblu}{RGB}{0,0,130}
\definecolor{dgre}{RGB}{0,130,0}
\definecolor{dgra}{RGB}{50,50,50}
\definecolor{mgra}{RGB}{100,100,100}
\definecolor{lgra}{RGB}{220,220,220}
\definecolor{MPG}{RGB}{000,125,122}
\definecolor{ora}{HTML}{FF9933}

\definecolor{AMPurple}{HTML}{663366}
\definecolor{Burgundy}{HTML}{993333}
\definecolor{Coffee}{HTML}{7B6049}
\definecolor{ForestGreen}{HTML}{005826}
\definecolor{Lavender}{HTML}{6E6AB1}
\definecolor{PSLightBlue}{HTML}{7DA7D9}

\newcommand{\g}{\,|\,}
\newcommand{\de}{\partial}
\newcommand{\Exp}{\mathbb{E}}
\newcommand{\cov}{\operatorname{cov}}
\newcommand{\diag}{\operatorname{diag}}
\newcommand{\N}{\mathcal{N}}
\newcommand{\T}{^{\intercal}}

\newcommand{\q}{\quad}
\newcommand{\qq}{\qquad}

\newcommand{\qqqq}{\qquad\qquad}

\renewcommand{\vec}{\boldsymbol}

\newcommand{\G}{\mathcal{G}}
\newcommand{\GP}{\mathcal{GP}}
\newcommand{\Id}{\vec{I}}

\newcommand{\tr}{\operatorname{tr}}

\newcommand{\blkdiag}{\operatorname{blkdiag}}

\newcommand{\w}{\vec{w}}

\newcommand{\y}{\vec{y}}
\newcommand{\x}{\vec{x}}

\usepackage{colonequals}
\newcommand{\ce}{\colonequals}

\renewcommand{\Re}{\mathbb{R}}

\usepackage{nicefrac}

\usetikzlibrary{arrows,shapes,calc}

\tikzset{>=stealth'}
\tikzstyle{graphnode} =
   [circle,draw=black,minimum size=22pt,text centered,text
     width=22pt,inner sep=0pt]
\tikzstyle{var}   =[graphnode,fill=white]
\tikzstyle{obs}   =[graphnode,fill=black,text=white]
\tikzstyle{fac}   =[rectangle,draw=black,fill=black!25,minimum size=5pt]
\tikzstyle{facprior} =[rectangle,draw=black,fill=black,text=white,minimum
size=5pt]
\tikzstyle{edge}  =[draw=white,double=black,thick,-]
\tikzstyle{prior} =[rectangle, draw=black, fill=black, minimum size=
5pt, inner sep=0pt]
\tikzstyle{dirprior} = [circle, draw=black, fill=black, minimum
size=5pt, inner sep=0pt]

\DeclareSymbolFont{stmry}{U}{stmry}{m}{n}
\DeclareMathSymbol\leftarrowtriangle\mathrel{stmry}{"5E}
\DeclareMathSymbol\rightarrowtriangle\mathrel{stmry}{"5F}

\renewcommand{\to}{\operatorname*{\rightarrowtriangle}}
\newcommand{\const}{\operatorname{const}}

\newif\iffinal 

\iffinal
 
\else
 
\fi

\pgfrealjobname{dual_control}

\pgfplotsset{compat=newest}

\newcommand{\inv}{^{-1}}
\renewcommand{\tr}{\operatorname{tr}}

\newcommand{\An}{\bar A}
\newcommand{\Bn}{\bar B}

\newcommand{\Kn}{\bar K}

\newcommand{\pn}{\bar p}

\newcommand{\un}{\bar u}

\newcommand{\zn}{\bar z}
\newcommand{\Jn}{\bar J}

\newcommand{\Ap}{\tilde A}
\newcommand{\Bp}{\tilde B}
\newcommand{\gp}{\tilde g}
\newcommand{\Kp}{\tilde K}
\newcommand{\pp}{\tilde p}

\newcommand{\Wp}{\tilde W}

\newcommand{\dx}{\Delta x}
\newcommand{\dz}{\Delta z}
\newcommand{\dhx}{\Delta \hat x}
\newcommand{\dhz}{\Delta \hat z}
\newcommand{\du}{\Delta u}
\newcommand{\dJ}{\Delta J}
\newcommand{\dt}{\Delta \t}

\newcommand{\sbk}{\sigma^2_{k}}
\newcommand{\sbkp}{\sigma^2_{k+1}}
\newcommand{\mbk}{\mu_{k}}
\newcommand{\mbkp}{\mu_{k+1}}

\newcommand{\uk}{u_k}

\renewcommand{\L}{\Lambda}
\newcommand{\cL}{\mathcal{L}}

\renewcommand{\t}{\theta}
\renewcommand{\tt}{\theta\theta}

\newcommand{\CE}{\textsc{ce}}
\newcommand{\OF}{\textsc{of}}
\newcommand{\BEB}{\textsc{beb}}
\newcommand{\DC}{\textsc{dc}}

\newcommand{\SE}{\textsc{se}}

\renewcommand{\P}{\mathcal{P}}
\newcommand{\Q}{\mathcal{Q}}
\newcommand{\R}{\mathcal{R}}
\newcommand{\K}{\mathcal{K}}
\newcommand{\F}{\mathcal{F}}
\renewcommand{\G}{\mathcal{G}}

\newcommand{\s}{\sigma}
\renewcommand{\w}{\omega}

\newcommand{\eg}{e.g.,~}
\newcommand{\ie}{i.e.~}

\newcommand{\wrt}{w.r.t.~}
\newcommand{\aka}{a.k.a.~}

\newlength\figureheight
\newlength\figurewidth

\begin{document}

\title{Dual Control\\ for Approximate Bayesian Reinforcement Learning}

\author{\name Edgar D.~Klenske \email edgar.klenske@tuebingen.mpg.de \\
        \addr Max-Planck-Institute for Intelligent Systems\\
              Spemannstra{\ss}e 38\\
              72076 T\"ubingen, Germany
        \AND
        \name Philipp Hennig \email philipp.hennig@tuebingen.mpg.de \\
        \addr Max-Planck-Institute for Intelligent Systems\\
              Spemannstra{\ss}e 38\\
              72076 T\"ubingen, Germany
}

\editor{Manfred Opper}

\maketitle

\begin{abstract}%
Control of non-episodic, finite-horizon dynamical systems with
uncertain dynamics poses a tough and elementary case of the
exploration-exploitation trade-off. Bayesian reinforcement learning, reasoning
about the effect of actions and future observations, offers a principled
solution, but is intractable. We review, then extend an old approximate
approach from control theory---where the problem is known as \emph{dual
control}---in the context of modern regression methods, specifically generalized
linear regression. Experiments on simulated systems show that this framework
offers a useful approximation to the intractable aspects of Bayesian RL,
producing structured exploration strategies that differ from standard RL
approaches. We provide simple examples for the use of this framework in
(approximate) Gaussian process regression and feedforward neural networks for
the control of exploration.
\end{abstract}


\section{Introduction}
\label{sec:introduction}

The exploration-exploitation trade-off is a central problem of learning in
interactive settings, where the learner's actions influence future
observations. In episodic settings, where the control problem is
re-instantiated repeatedly with unchanged dynamics, comparably simple notions of
exploration can succeed. E.g.,\ assigning an \emph{exploration bonus} to
uncertain options \citep{macready1998bandit,audibert2009exploration} or acting
optimally under one sample from the current probabilistic model of the
environment \citep[\emph{Thompson sampling}, see][]{Thompson,
chapelle2011empirical}, can perform well \citep{DeardenFA99Modelbased,
kolter2009near,SrinivasKKS2010Gaussian}. Such approaches, however, do not model
the effect of actions on future beliefs, which limits the potential for the
balancing of exploration and exploitation. This issue is most drastic in the
non-episodic case, the control of a single, ongoing trial. Here, the
controller cannot hope to be returned to known states, and exploration must be
carefully controlled to avoid disaster.

A principled solution to this problem is offered by \emph{Bayesian
reinforcement learning} \citep{duff2002optimal, poupart2006analytic,
GaussianRL}: A probabilistic belief over the dynamics and cost of the
environment can be used not just to simulate and plan trajectories, but also to
reason about changes to the belief from future observations, and their
influence on future decisions. An elegant formulation is to combine the physical
state with the parameters of the probabilistic model into an augmented
dynamical description, then aim to control this system. Due to the inference,
the augmented system invariably has strongly nonlinear dynamics, causing
prohibitive computational cost---even for finite state spaces and discrete time
\citep{poupart2006analytic}, all the more for continuous space and time
\citep{GaussianRL}.

The idea of augmenting the physical state with model parameters was noted
early, and termed \emph{dual control}, by \citet{Feldbaum1960Dual}. It
seems both conceptual and---by the standards of the time---computational
complexity hindered its application. An exception is a strand of several works
by Meier, Bar-Shalom, and Tse \citep{tse1973wide, tse1973actively,
bar1976caution, bar1981stochastic}. These authors developed techniques for
limiting the computational cost of dual control that, from a modern
perspective, can be seen as a form of approximate inference for Bayesian
reinforcement learning. While the Bayesian reinforcement learning community is
certainly aware of their work \citep{duff2002optimal, GaussianRL}, it has not
found widespread attention. The first purpose of this paper is to cast their
dual control algorithm as an approximate inference technique for Bayesian RL in
parametric Gaussian (general least-squares) regression. We then extend the
framework with ideas from contemporary machine learning. Specifically, we
explain how it can in principle be formulated non-parametrically in a Gaussian
process context, and then investigate simple, practical finite-dimensional
approximations to this result. We also give a simple, small-scale example for
the use of this algorithm for dual control if the environment model is
constructed with a feedforward neural network rather than a Gaussian process.

\section{Model and Notation}
\label{sec:model}

Throughout, we consider discrete-time, finite-horizon dynamic systems (POMDPs)
of form
\begin{equation*} \label{eq:system}
  x_{k+1} = f_k(x_k, u_k) + \xi_k \q\text{(state dynamics)} \qqqq
  y_{k} = Cx_k + \gamma_k \q \text{(observation model)}.
\end{equation*}
At time $k\in\{0,\dots,T\}$, $x_k\in\Re^n$ is the state, $\xi_k\sim\N(0,Q)$ is a
Gaussian disturbance. The control input (continuous action) is denoted $u_k$;
for simplicity we will assume scalar $u_k\in\Re$ throughout. Measurements
$y_k\in\Re^d$ are observations of $x_k$, corrupted by Gaussian noise
$\gamma_k\sim\N(0,R)$. The generative model thus reads $p(x_{k+1}\g x_k,u_k) =
\N(x_{k+1};f_k(x_k,u_k),Q)$ and $p(y_k\g x_k)=\N(y_k;Cx_k,R)$, with a linear map
$C\in\Re^{d\times
  n}$. Trajectories are vectors $\vec{x}=[x_0,\dots,x_T]$, and analogously for
$\vec{u},\vec{y}$. We will occasionally use the subset notation
$\vec{y}_{i:j}=[y_i,\dots,y_j]$. We further assume that dynamics $f_k$ are not
known, but can be described up to Gaussian uncertainty by a general linear model
with nonlinear features $\phi:\Re^n\to\Re^{m}$ and uncertain matrices $A_k,B_k$.
 \begin{equation}
  \label{eq:feature-linear-system}
  x_{k+1} = A_k\phi(x_k) + B_k u_k + \xi_k, \qqqq A_k\in\Re^{n\times m};
  B_k\in\Re^{n\times 1}.
\end{equation}
To simplify notation, we reshape the elements of $A_k$ and $B_k$ into a
parameter vector
$\t_k=[\operatorname{vec}(A_k);\operatorname{vec}(B_k)]\in\Re^{(m+1)n}$,
and define the reshaping transformations $A(\t_k): \t_k \mapsto A_k$ and
$B(\t_k): \t_k \mapsto
B_k$.  At initialization, $k=0$, the belief over states and parameters
is assumed to be Gaussian
\begin{equation}
\label{eq:13}
  p\left(\begin{bmatrix}x_0\\ \t_0 \end{bmatrix}\right)
  = \N\left(
  \begin{bmatrix}
    x_0\\ \t_0
  \end{bmatrix}
  ;
  \begin{bmatrix}
    \hat x_0\\ \hat{\t}_0
  \end{bmatrix}
  ,
  \begin{bmatrix}
    \Sigma^{xx} _0 & \Sigma^{x\t} _0 \\ \Sigma^{\t x} _0 & \Sigma^{\t\t} _0
  \end{bmatrix} \right).
\end{equation}
The control response $B_k u_k$ is linear, a
common assumption for physical systems. Nonlinear mappings can be included in a
generic form $\phi(x_k, u_k)$, but complicate the following derivations and
raise issues of identifiability. For simplicity, we also assume that the
dynamics do not change through time:
$p(\t_{k+1}\g\t_k)=\delta(\t_{k+1}-\t_{k})$. This could be relaxed to an
autoregressive model $p(\t_{k+1}\g\t_k)=\N(\t_{k+1};D\t_{k},\Xi)$, which would
give additive terms in the derivations below. Throughout, we assume a finite
horizon with terminal time $T$ and a quadratic cost function in state and
control
\begin{equation*}
  \label{eq:cost}
  \cL(\vec{x},\vec{u}) = \left[\sum_{k=0}^{T} (x_k - r_k)\T W_k (x_k
    - r_k) +
    \sum_{k=0}^{T-1} u_k\T U_k u_k \right],
\end{equation*}
where $\vec{r} = [r_0,\dots,r_T]$ is a target trajectory.  $W_k$ and $U_k$
define state and control cost, they can be time-varying. The goal, in line with
the standard in both optimal control and reinforcement learning, is to find the
control sequence $\vec{u}$ that, at each $k$, minimizes the \emph{expected cost}
to the horizon
\begin{equation}
  \label{eq:1}
  J_k(\vec{u}_{k:T-1},p(x_k)) = \Exp_{x_k}\left[ (x_k - r_k)\T W_k (x_k -
r_k) + u_k\T U_k u_k +
J_{k+1}(\vec{u}_{k+1:T-1},p(x_{k+1}))\g p(x_k)\right],
\end{equation}
where past measurements $\vec{y}_{1:k}$, controls $\vec{u}_{1:k-1}$ and
prior information $p(x_0)$ are incorporated into the belief $p(x_k)$, relative
to which the expectation is calculated. Effectively, $p(x_k)$ serves as a
bounded rationality approximation to the true information state. Since the
equation above is recursive, the final element of the cost has to be defined
differently, as
\begin{equation*}
    J_{T}(p(x_T)) = \Exp_{x_{T}}\left[ (x_T - r_T)\T W_T (x_T - r_T)
  \g p(x_T) \right]
\end{equation*}
(that is, without control input and future cost). The optimal control sequence
minimizing this cost will be denoted $\vec{u}^*$, with associated cost
\begin{equation}
  \label{eq:optimal-cost}
  J^*_k(p(x_k)) = \min_{u_k}\Exp_{x_k}\left[ (x_k - r_k)\T W_k (x_k - r_k) +
    u_k\T U_k u_k + J_{k+1} ^*(p(x_{k+1}))\g p(x_k) \right].
\end{equation}
This recursive formulation, if written out, amounts to alternating minimization
and expectation steps. As $u_k$ influences $x_{k+1}$ and $y_{k+1}$, it enters
the latter expectation nonlinearly. Classic optimal control is the linear base
case ($\phi(x)=x$) with known $\t$, where $\vec{u}^*$ can be found by dynamic
programming \citep{Bellman1961Adaptive,Bertsekas2005Dynamic}.

\section{Bayesian RL and Dual Control}
\label{sec:bayesian-rl}

\citet{Feldbaum1960Dual} coined the term \emph{dual control} to describe the
idea now also known as Bayesian reinforcement learning in the machine learning
community: While adaptive control only considers past observations, dual
control also takes future observations into account. This is necessary because
all other ways to deal with uncertain parameters have substantial drawbacks.
Robust controllers, for example, sacrifice performance due to their
conservative design; adaptive controllers based on \emph{certainty
equivalence} (where the uncertainty of the parameters is not taken into account
but only their mean estimates) do not show exploration, so that all learning is
purely passive. For most systems it is obvious that more excitation leads to
better estimation, but also to worse control performance. Attempts at finding a
compromise between exploration and exploitation are generally subsumed under the
term ``dual control'' in the control literature. It can only be achieved by
taking the future effect of current actions into account.

It has been shown that optimal dual control is practically unsolvable for most
cases \citep{Aoki1967Optimization}, with a few examples where solutions were
found for simple systems \cite[\eg][]{Sternby1976Simple}. Instead, a large
number of \emph{approximate} formulations of the dual control problem were
formulated in the decades since then.  This includes the introduction of
perturbation signals \cite[\eg][]{JacobsP1972Caution}, constrained optimization
to limit the minimal control signal or the maximum variance, serial expansion
of the loss function \cite[\eg][]{tse1973wide} or modifications of the loss
function \cite[\eg][]{FilatovUnbehauen2004Adaptive}. A comprehensive overview
of dual control methods is given by \cite{Wittenmark1995Adaptive}. A historical
side-effect of these numerous treatments is that the meaning of the term ``dual
control'' has evolved over time, and is now applied both to the fundamental
concept of optimal exploration, and to methods that only approximate this
notion to varying degree. Our treatment below studies one such class of
practical methods that aim to approximate the true dual control solution.

The central observation in Bayesian RL / dual control is that both the
states $x$ and the parameters $\t$ are subject to uncertainty.
While part of this uncertainty is caused by randomness, part by lack of
knowledge, both can be captured in the same way by probability distributions.
States and parameters can thus be subsumed in an \emph{augmented state}
\citep{Feldbaum1960Dual,duff2002optimal,poupart2006analytic} $z_k\T =
\begin{pmatrix} x_k\T & \t_k\T\end{pmatrix}\in\Re^{(m+2)n}$. In this
notation, the optimal exploration-exploitation
trade-off---relative to the probabilistic priors defined above---can be written
compactly as optimal control of the augmented system with a new observation
model $p(y_k\g z_k) =\N(y_k;\tilde{C}z_k,R)$ using
$\tilde{C}=\begin{bmatrix}C & \vec{0}\end{bmatrix}$ and
a cost analogous to Eq.~(\ref{eq:1}).

Unfortunately, the dynamics of this new system are nonlinear, even if the
original physical system is linear. This is because inference is always
nonlinear and future states influence future parameter beliefs, and vice versa.
A first problem, not unique to dual control, is thus that inference is not
analytically tractable, even under the Gaussian assumptions above
\citep{Aoki1967Optimization}. The standard remedy is to use approximations,
most popularly the linearization of the extended Kalman filter
\citep[e.g.,][]{sarkka2013bayesian}. This gives a sequence of approximate
Gaussian likelihood terms. But even so, incorporating these Gaussian likelihood
terms into future dynamics is still intractable, because it involves
expectations over rational polynomial functions, whose degree increases with
the length of the prediction horizon. The following section provides an
intuition for this complexity, but also the descriptive power of the augmented
state space.

As an aside, we note that several authors \citep{Kappen2011Optimal, GaussianRL}
have previously pointed out another possible construction of an augmented state:
incorporating not the actual \emph{value} of the parameters $\theta_k$ in the
state, but the parameters $\mu_k$, $\Sigma_k$ of a Gaussian belief
$p(\theta_k\g \mu_k,\Sigma_k) = \N(\theta_k;\mu_k,\Sigma_k)$ over them. The
advantage of this is that, if the state $x_k$ is observed without noise, these
belief parameters follow stochastic differential equations---more
precisely, $\Sigma_k$ follows an ordinary (deterministic) differential equation,
while $\mu_k$ follows a stochastic differential equation---and it can then be
attempted to solve the control problem for these differential equations more
directly.

While it can be a numerical advantage, this formulation of the augmented state
also has some drawbacks, which is why we have here decided not to adopt it:
First, the simplicity of the directly formalizable SDE vanishes in the POMDP
setting, \ie if the state is not observed without noise. If the state
observations are corrupted, the exact belief state is not a Gaussian process,
so that the parameters $\mu_k$ and $\Sigma_k$ have no natural meaning.
Approximate methods can be used to retain a Gaussian belief (and we will do so
below), but the dynamics of $\mu_k$, $\Sigma_k$ are then intertwined with the
chosen approximation (\ie changing the approximation changes their dynamics),
which causes additional complication. More generally speaking, it is not
entirely natural to give differing treatment to the state $x_k$ and parameters
$\theta_k$: Both state and parameters should thus be treated within the same
framework; this also allows extending the framework to the case where also the
parameters do follow an SDE.

\subsection{A Toy Problem}
\label{sec:toy-problem}
To provide an intuition for sheer complexity of optimal dual control, consider
the perhaps simplest possible example: the linear, scalar system
\begin{equation}
  \label{eq:toy-example}
  x_{k+1} = a x_k + b u_k + \xi_k,
\end{equation}
with target $r_k=0$ and noise-free observations ($R=0$). If $a$ and $b$ are
known, the optimal $u_k$ to drive the current state $x_k$ to zero in one step
can be trivially verified to be
\begin{equation*}
  u_{k,\text{oracle}} ^* = -\frac{a b x_k}{U + b^2}.
\end{equation*}

Let now parameter $b$ be uncertain, with current belief $p(b) = \N(b; \mbk,
\sbk)$ at time $k$. The na\"ive option of simply replacing the parameter with
the current mean estimate is known as \emph{certainty equivalence (CE)} control
in the dual control literature \citep[\eg][]{BarShalomT1974Dual}. The resulting
control law is
\begin{equation*}
  u_{k,\CE}^* = -\frac{a\mbk x_k}{U + \mbk^2}.
\end{equation*}
It is used in many adaptive control settings in practice, but has substantial
deficiencies: If the uncertainty is large, the mean is not a good estimate, and
the CE controller might apply completely useless control signals. This often
results in large overshoots at the beginning or after parameter changes.

A slightly more elaborate solution is to compute the expected cost $\Exp_{b}[
x_{k+1}^2 + U u_k ^2 \g \mbk, \sbk]$ and then optimize for $u_k$. This gives
\emph{optimal
feedback (OF)} or ``cautious'' control \citep{Dreyfus64Some}\footnote{Dreyfus
used the term ``open loop optimal feedback'' for his approach, a term that is
misleading to modern readers, because it is in fact a closed-loop algorithm.}:
\begin{equation}\label{eq:2}
  u_{k,\OF}^* = -\frac{a\mbk x_k}{U + \sbk + \mbk^2}.
\end{equation}
This control law reduces control actions in cases of high parameter
uncertainty. This mitigates the main drawback of the CE controller, but leads to
another problem: Since the OF controller decreases control with rising
uncertainty, it can entirely prevent learning. Consider the posterior on
$b$ after observing $x_{k+1}$, which is a closed-form Gaussian (because
$u_k$ is chosen by the controller and has no uncertainty):
\begin{equation}\label{eq:belief-update}
  p(b\g \mbkp, \sbkp) = \N(b; \mu_{k+1}, \sbkp) \\ = \N\left(b;
  \frac{\sbk \uk (b \uk + \xi_k) + \mbk Q}{\uk^2 \sbk + Q}, \frac{\sbk
  Q}{\uk^2\sbk + Q}\right)
\end{equation}
($b$ shows up in the fully observed $x_{k+1}=ax_k + bu_k+\xi_k$). The dual
effect here is that the updated $\sbkp$ depends on $\uk$. For large values of
$\sigma_k ^2$, according to \eqref{eq:2},  $u_{k,\OF} ^*\to 0$, and the new
uncertainty $\sigma_{k+1} ^2\to \sigma_{k} ^2$. The system thus will never learn
or act, even for large $x_k$. This is known as the ``turn-off phenomenon''
\citep{Aoki1967Optimization, bar1981stochastic}.

However, the derivation for OF control above amounts to minimizing
Eq.~\eqref{eq:1} for the myopic controller, where the horizon is only a single
step long ($T=1$). Therefore, OF control is indeed optimal for this case. By
the optimality principle \cite[e.g.,][]{Bertsekas2005Dynamic}, this means that
Eq.~\eqref{eq:2} is the optimal solution for the last step of every controller.
But since it does not show any form of exploration or ``probing''
\citep{bar1976caution}, a myopic controller is not enough to show the dual
properties.

In order to expose the dual features, the horizon has to be at least of length
$T=2$. Since the optimal controller follows Bellman's equation, the solution
proceeds backwards. The solution for the second control action $u_1$ is
identical to the solution of the myopic controller \eqref{eq:2}; but after
applying the first control action $u_0$, the belief over the unknown parameter
$b$ needs an update according to Eq.~\eqref{eq:belief-update}, resulting in
\begin{equation}
  \label{eq:preposterior-control}
  u_{1}^* = -\left[U +
    \frac{\sigma_0 ^2 Q}{u_0 ^2\sigma_0 ^2
      +  Q} + \left(\frac{\sigma_0 ^2 u_0 (b u_0 + \xi_0) + \mu_0
        Q}{u_0 ^2 \sigma_0 ^2 + Q}\right)^2\right]^{-1}
    \left[a \frac{\sigma_0 ^2 u_0 (b u_0 +
      \xi_0) + \mu_0 Q}{u_0 ^2 \sigma_0 ^2 + Q} x_{1}\right].
\end{equation}

\begin{figure}
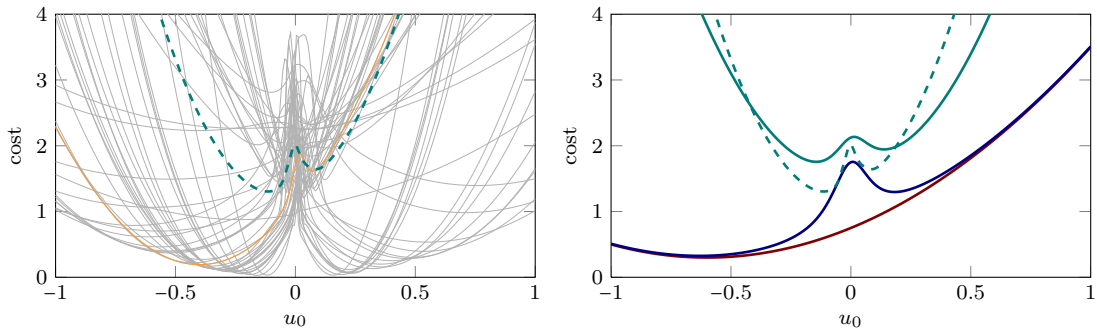

  \setlength\figureheight{3.5cm}
  \setlength\figurewidth{0.44\columnwidth}
  \scriptsize
  \mbox{
   \beginpgfgraphicnamed{dual_control_sampled_b-external}%
   \input{dual_control_sampled_b.tikz}%
   \endpgfgraphicnamed%

   \beginpgfgraphicnamed{dual_control_uncertain_b-external}%
   \input{dual_control_uncertain_b.tikz}%
   \endpgfgraphicnamed%
 
  }
  \caption{{\bf Left:} Computing the $T=2$ dual cost for the simple
    system of Eq.~\eqref{eq:toy-example}. Costs $\cL(u_0)$ under
    optimal control on $u_1$ for sampled parameter $b$ (thin gray;
    one sample highlighted, orange). Expected dual cost $J(u_0)$
    under $u_1 ^*$ (dashed green). The optimal $u_0 ^*$ lies at the
    minimum of the dashed green line. {\bf Right:} Comparison of sampling
    (dashed green; thin gray: samples) to three approximations: CE (red) and CE
with Bayesian exploration bonus (blue). The solid green line is the approximate
dual control
    constructed in Section \ref{sec:appr-dual-contr}. See also
Sec.~\ref{sec:exp-simple-scalar} for details.}
  \label{fig:sampling_uncertain_b}
\end{figure}

Inserting into Eq.~(\ref{eq:optimal-cost}) gives
\begin{equation}
  \label{eq:3}
  \begin{split}
    J_0^*(x_0) &= \min_{u_0} \Exp_{x_0}\left[W x_0^2 + U
      u_0^2 + \min_{u_1} \Exp_{x_1} \left[ Wx_1^2 + U u_1 ^2
        +\Exp_{x_2}[Wx_2 ^2]\right]\right] \\
    &= \min_{u_0} \left[ W x_0^2 + U u_0^2 +
      \Exp_{\xi_0,b} \left[ Wx_1^2 + U (u_1 ^*) ^2
        +\Exp_{\xi_1,b}\left[W(x_1+b u_1 ^* + \xi_1)^2\g \mu_1, \sigma_1 \right]
    \g \mu_0, \sigma_0\right]\right].
  \end{split}
\end{equation}
Since $u_1 ^*$ from Eq.~(\ref{eq:preposterior-control}) is already a rational
function of fourth order in $b_0$, and shows up quadratically in
Eq.~(\ref{eq:3}), the relevant expectations cannot be computed in closed form
\cite[][]{Aoki1967Optimization}.  For this simple case though, it is possible to
compute the optimal dual control by performing the expectation through sampling
$b, \xi_0, \xi_{1}$ from the prior. Fig.~\ref{fig:sampling_uncertain_b} shows
such samples of $\cL(u_0)$ (in gray; one single sample highlighted in orange),
and the empirical expectation $J(u_0)$ in dashed green. Each sample is a
rational function of even leading order. In contrast to the CE cost,
the dual cost is much narrower, leading to more cautious behavior of the dual
controller. The average dual cost has its minima not at zero, but to either
side of it, reflecting the optimal amount of exploration in this
particular belief state.

While it is not out of the question that the Monte Carlo solution can remain
feasible for larger horizons, we are not aware of successful solutions for
continuous state spaces \citep[however, see][for a sampling solution to
Bayesian reinforcement learning in discrete spaces, including notes on the
considerable computational complexity of this approach]{poupart2006analytic}.
The next section describes a tractable \emph{analytic} approximation that does
not involve samples.

\section{Approximate Dual Control for Linear Systems}
\label{sec:appr-dual-contr}

\begin{figure}
  \footnotesize
  \centering
   \beginpgfgraphicnamed{the_algorithm-external}%
\tikzstyle{block} = [draw, fill=black!5, rectangle,
    minimum height=0.5cm, minimum width=2.5cm, align=center]
\tikzstyle{decision} = [draw, fill=black!5, diamond, aspect=2,
    minimum height=0.5cm, minimum width=2.5cm, align=center]

\begin{tikzpicture}[auto, node distance=1.5cm,>=open triangle 45]
    \node[block] (init) {Initialize};
    \node [block, right of= init, node distance=3cm] (compute_CE)
{Compute \\ {\bfseries CE control}};
	\node [block, below of= compute_CE]
(extrapolate) { {\bfseries predict} state \\and
covariance\\ for given $u_k$};
	\node [block, below of= extrapolate, node distance=1.65cm] (nominal)
{\ding{172} Compute \\
{\bfseries
CE trajectory}\\ and its covariances};
	\node [block, below of= nominal] (dual_cost) {\ding{173} Evaluate\\
\bfseries{cost-to-go}};
  \node [decision, below of= dual_cost, minimum height=0.2cm, inner sep=0, node
distance=1.5cm] (stop) {search
over?};
  \node [block, left of= stop, node distance=4.5cm] (next_uk) {\ding{174}
Compute next
value \\${u_k}$
for the {\bfseries search}};
  \node [block, right of= stop, node distance=4cm] (control) {Apply the \\
{\bfseries control}};
  \node [block, right of= dual_cost, node distance=4cm] (simulate) {Simulate
or {\bfseries run}
\\the system};
  \node [block, right of= nominal, node distance=4cm] (measurement) {Make new
\\ {\bfseries
measurement}};

  \draw [->] (init) -- (compute_CE);
  \draw [->] (compute_CE) -- (extrapolate);
  \draw [->] (extrapolate) -- (nominal);
  \draw [->] (nominal) -- (dual_cost);
  \draw [->] (dual_cost) -- (stop);
  \draw [->] (stop) -- node[above]{yes} (control);
  \draw [->] (stop) -- node[above]{no} (next_uk);
  \draw [->] (next_uk) |- (extrapolate);
  \draw [->] (control) --(simulate);
  \draw [->] (simulate) --(measurement);
  \draw [->] (measurement) |- (compute_CE);
\end{tikzpicture}%
   \endpgfgraphicnamed%
 \vspace{-2mm}\\
  \caption{
  Flow-chart of the approximate dual control algorithm to show the overall
structure. Adapted from \cite{tse1973actively}. The left cycle is the inner
loop, performing the nonlinear optimization.}
  \label{fig:the-algorithm}
\end{figure}
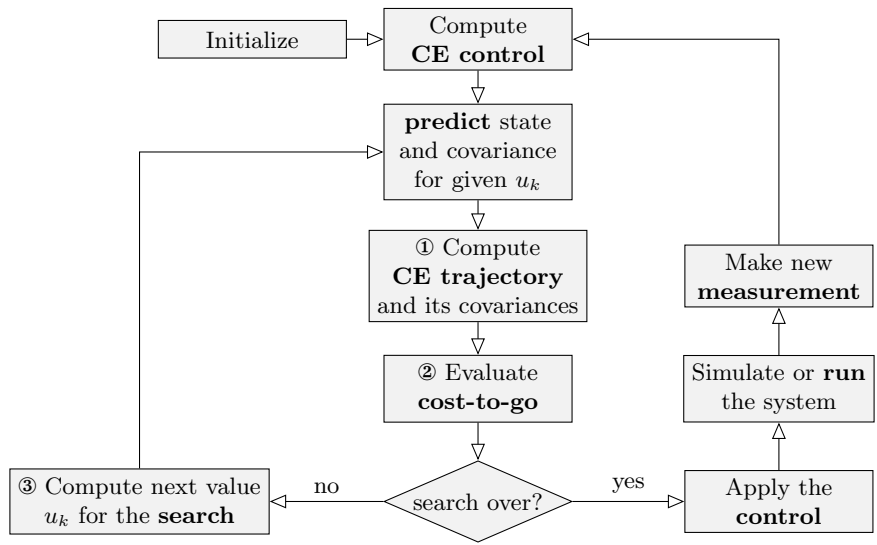

In 1973, \citet{tse1973wide} constructed theory and an
algorithm \citep{tse1973actively} for approximate dual (AD) control, based on
the series expansion of the cost-to-go. This is related to differential
dynamic programming for the control of nonlinear dynamic systems
\citep{Mayne1966Second}. It separates into three conceptual steps (described in
Sec.~\ref{sec:cert-equiv-contr}--\ref{sec:optim-curr-contr}), which together
yield what, from a contemporary perspective, amounts to a structured Gaussian
approximation to Bayesian RL:
\begin{dingautolist}{172}
\item Find an optimal trajectory for the deterministic part of the
system under the mean model: the \emph{nominal} trajectory under certainty
equivalent control. For linear systems this is easy (see below), for nonlinear
ones it poses a nontrivial, but feasible nonlinear model predictive control
problem \citep{allgower1999nonlinear, diehl2009nonlinear}. It yields a nominal
trajectory, relative to which the following step constructs a
tractable quadratic expansion.
 \item Around the nominal trajectory, construct a local \emph{quadratic
     expansion} that approximates the effects of future
  observations. Because the expansion is quadratic, an optimal control
  law relative to the deterministic system---the \emph{perturbation
    control}---can be constructed by dynamic programming. Plugging
  this perturbation control into the residual dynamics of the
  approximate quadratic system gives an approximation for the
  cost-to-go.
  This step adds the cost of uncertainty to the deterministic control cost.
\item In the current time step $k$, perform the prediction for an
  arbitrary control input $u_k$ (as opposed to the analytically
  computed control input for later steps). Optimize $u_k$ numerically by
  repeated computation of steps \ding{172} and \ding{173} at
  varying $u_k$ to minimize the approximate cost.
\end{dingautolist}
These three steps will be explained in detail in the subsequent sections.
The interplay between the different parts of the algorithm is shown in
Figure~\ref{fig:the-algorithm}.

The abstract introductory work \cite{tse1973wide} is relatively general, but the
explicit formulation in \cite{tse1973actively} only applies to linear systems.
Since both works are difficult to parse for contemporary readers, the following
sections thus first provide a short review, before we extend to more
modern concepts. In this section, we follow the more transparent case of a
linear system from \cite{tse1973actively}, \ie $\phi(x)=x$ in
Eq.~(\ref{eq:feature-linear-system}). For the augmented state $z$, this still
gives a nonlinear system, because $\theta$ and $x$ interact multiplicatively
\begin{equation}
  \label{eq:augmented-system}
  z_{k+1} =
  \begin{pmatrix}
    x_{k+1}\\\theta_{k+1}
  \end{pmatrix} =
  \begin{pmatrix}
    A(\theta_k) & 0 \\ 0 & \Id
  \end{pmatrix}z_k +
  \begin{pmatrix}
    B(\theta_k) \\ 0
  \end{pmatrix}u_k +
  \begin{pmatrix}
    \xi_k \\ 0
  \end{pmatrix}
  \equalscolon
  \tilde f(z_k, u_k).
\end{equation}
The parameters $\t$ are assumed to be
deterministic, but not known to the controller. This uncertainty is captured by
the distribution $p(\t)$ representing the lack of knowledge.

\subsection{Certainty Equivalent Control Gives a Nominal Reference
  Trajectory}
\label{sec:cert-equiv-contr}

The certainty equivalent model is built on the assumption that the uncertain
$\theta$ coincide with their most likely value, the mean
$\hat\t$ of $p(\theta)$, and that the system propagates
deterministically without noise.
This means that the nominal parameters $\bar\t$ are the current mean values
$\hat\t$, which decouples $\theta$ entirely
from $x$ in Eq.~(\ref{eq:augmented-system}), and the optimal control for the
finite horizon
problem can be computed by dynamic programming (DP)
\citep{Aoki1967Optimization}, yielding an optimal linear control law
\begin{equation*}
  \un_j^* = -\left(\Bn\T \Kn_{j+1} \Bn +
    U_j\right)\inv \Bn\T \left[ \Kn_{j+1} \An \bar{x}_j + \pn_{j+1} \right],
\end{equation*}
where we have momentarily simplified notation to
$\An=A(\bar{\theta}_j),\Bn=B(\bar{\theta}_j),\,\forall j$, because the
$\bar{\theta}_j$  are constant. The $\Kn_{j}$ and $\pn_{j}$ for $j=k+1,\dots,T$
are defined and computed recursively as
\begin{subequations}
  \label{eq:dp-nominal}
  \begin{xalignat*}{2}
    \Kn_j &= \An\T\left(\Kn_{j+1}-\Kn_{j+1}\Bn\left(\Bn\T
        \Kn_{j+1} \Bn
        + U_j\right)\inv \Bn\T \Kn_{j+1} \right) \An + W_j & \Kn_T &= W_T\\
    \pn_j &= \An\T\left(\pn_{j+1}-\Kn_{j+1}\Bn\left(\Bn\T
        \Kn_{j+1} \Bn + U_j\right)\inv \Bn\T \Kn_{j+1} \right)
    \pn_{j+1} - W_j r_j & \pn_T &= -W_T r_T,
  \end{xalignat*}
\end{subequations}
where $\vec{r}$ is the reference trajectory to be followed.
This CE controller gives the \emph{nominal trajectory} of inputs
$\bar{\vec{u}}_{k:T-1}$ and states $\bar{\vec{x}}_{k:T}$, from the current time
$k$ to the horizon $T$. The true future trajectory is subject to stochasticity
and uncertainty, but the deterministic nominal trajectory $\bar{\vec{x}}$, with
its optimal control $\bar{\vec{u}}^*$ and associated nominal cost $\bar{J}^*_k
= \cL(\bar{\vec{x}}_{k:T},\bar{\vec{u}}^* _{k:T})$ provides a base, relative to
which an approximation will be constructed.

\subsection{Quadratic Expansion Around the Nominal Defines Cost of Uncertainty}
\label{sec:quadr-expans-around}

The central idea of AD control is to project the nonlinear objective
$J_k(\vec{u}_{k:T-1}, p(x_k))$ of Eq.~(\ref{eq:1}) into a quadratic, by
locally linearizing around the nominal trajectory $\vec{x}$ and maintaining a
joint Gaussian belief.

%
%
%

To do so, we introduce small perturbations around nominal cost, states, and
control: $\dJ_j = J_j - \Jn_j, \dz_j = z_j - \zn_j,$ and $\du_j = u_j - \un_j$.
These perturbations arise from both the stochasticity of the state and the
parameter uncertainty. Note that a change in the state results in a change of
the control signal, because the optimal control signal in each step depends on
the state. Even though the origin of the uncertainties is different ($\dx$
arises from stochasticity and $\dt$ from the lack of knowledge), both can be
modeled in a joint probability distribution.

Approximate Gaussian filtering ensures that beliefs over $\dz$ remain Gaussian:
\begin{equation*}
  \label{eq:6}
  p(\dz_j) = \N\left[
    \begin{pmatrix}
      \dx_j\\ \Delta \theta_j
    \end{pmatrix};
    \begin{pmatrix}
      \Delta \hat{x}_j \\ 0
    \end{pmatrix},
    \begin{pmatrix}
      \Sigma^{xx} _j & \Sigma^{x\theta} _j \\
      \Sigma^{\theta x} _j & \Sigma^{\theta\theta} _j \\
    \end{pmatrix} \right].
\end{equation*}
Note that shifting the mean to the nominal trajectory does not change the
uncertainty. Note further that the expected perturbation in the parameters is
nil. This is because the parameters are assumed to be deterministic and are not
affected by any state or input.

Calculating the Gaussian filtering updates is in principle not possible for
future measurements, since it violates the causality principle
\citep{GladLjung2000Control}. Nonetheless, it is possible to use the
\emph{expected} measurements to simulate the effects of the future measurements
on the uncertainty, since these effects are deterministic. This is sometimes
referred to as preposterior analysis \citep{RaiffaSchlaifer1961Applied}.

To second order around the nominal trajectory, the cost is approximated by
\begin{equation*}
    J_k(\vec{u}_{k:T-1}, p(x_{k})) = \Jn_k^* + \dJ_k
    \approx \Jn_k^* + \Delta \tilde J_k,
\end{equation*}
where $\bar J^*_k$ is the optimal cost for the nominal system and
$\Delta\tilde{J}_k$ is the approximate additional cost from the perturbation:
\begin{equation}
  \label{eq:5}
    \Delta \tilde J_k \ce \Exp_{\vec{x}_{k:T}}\left[ \sum_{j=k}^{T}
\left\{ (\bar{x}_j - r_j)\T W_j \dx_j + \frac{1}{2}\dx_j\T W_j \dx_j
    \right\} + \sum_{j=k}^{T-1} \left\{ \bar{u}_j\T
      U_j \du_j + \frac{1}{2}\du_j\T U_j \du_j\right\}\right].
\end{equation}
Although the uncertain parameters $\theta$ do not show up explicitly in the
above equation, this step captures dual effects: The uncertainty of the
trajectory $\Delta \vec{x}$ depends on $\theta$ via the dynamics. Higher
uncertainty over $\theta$ at time $j-1$ causes higher predictive uncertainty
over $\Delta x_{j}$ (for each $j$), and thus increases the expectation of the
quadratic term $\Delta x_j\T W_j \Delta x_j$. Control that decreases uncertainty
in $\theta$ can lower this approximate cost, modeling the benefit of
exploration. For the same reason, Eq.~\eqref{eq:5} is
in fact still not a quadratic function and has no closed form
solution. To make it tractable, \citet{tse1973actively} make the
ansatz that all terms in the expectation of Eq.~(\ref{eq:5}) can be written as
$g_j + p_j\T \Delta z_j + \nicefrac{1}{2}\Delta z_j\T K_j \Delta z_j$.
This amounts to applying dynamic programming on the perturbed system.
Expectations over the cost under Gaussian beliefs on $\Delta z$
can then be computed analytically. Because all $\Delta\theta$ have zero mean,
linear terms in these quantities vanish in the expectation.
This allows analytic minimization of the approximate optimal cost for each
time step
\begin{multline}
  \label{eq:optimal-perturbation-cost}
  \Delta \tilde J_j^*(p(x_j)) = \underset{\du_j}{\min} \left\{ (x_j -
  r_j)\T W_j
  \dhx_{j|j} + \frac{1}{2}\dhx_{j|j}\T W_j \dhx_{j|j} + u_j\T U \du_j +
  \frac{1}{2}\du_j\T U \du_j
    \right. \\ \left.
    + \frac{1}{2} \tr\left[W_j \Sigma^{xx}_{j|j}\right]
    + \underset{\dx_{j+1}}{\Exp} \left[
  \Delta\tilde J_{j+1}^*(\vec{y}_{1:j+1}) \g
  p(x_j) \right] \right\},
\end{multline}
which is feasible given an explicit description of the Gaussian filtering
update. It is important to note that, assuming extended Kalman filtering, the
update to the mean from \emph{expected} future observations $y_{j+1}$ is nil.
This is because we expect to see measurements consistent with the current mean
estimate. Nonetheless, the (co-)variance changes depending on the control input
$u_j$, which is the dual effect.

Following the dynamic programming equations for the perturbed problem,
including the additional cost from uncertainty, the resulting cost amounts to
\citep{tse1973wide}
\begin{multline*}
  \Delta\tilde{J}^*_k(p(x_k)) = \gp_{k+1}  + \pp_{k+1}\T\Delta\hat z_k +
  \frac{1}{2}  \Delta\hat z_k\T \Kp_{k+1}\Delta\hat z_k \\
    + \frac{1}{2} \tr\left\{ W_T\Sigma^{xx}_{T|T} + \sum_{j=k}^{T-1} \left[ W_j
    \Sigma^{xx}_{j|j} + (\Sigma_{j+1|j} - \Sigma_{j+1|j+1} ) \Kp_{j+1} \right]
\right\}
\end{multline*}
(where we have neglected second-order effects of the dynamics). Recalling that
$\dhz=0$ and dropping the constant part, the dual cost can be approximated to be
\begin{equation*}
  J^d_k = \frac{1}{2} \tr\left\{ W_T\Sigma^{xx}_{T|T} + \sum_{j=k}^{T-1}
    \left[ W_j \Sigma^{xx}_{j|j} + (\Sigma_{j+1|j} - \Sigma_{j+1|j+1} )
\Kp_{j+1}    \right] \right\} \qq \left(=\Delta\tilde J_k^* - \const\right)
\end{equation*}
where the recursive equation
\begin{equation*}
  \label{eq:dp-perturbed}
    \Kp_j = \Ap\T\left(\Kp_{j+1}-\Kp_{j+1}\Bp\left(B\T K^{xx}_{j+1} B
        + U_j\right)\inv \Bp\T \Kp_{j+1} \right) \Ap + \Wp_j \qqqq \Kp_T =
      \Wp_T
\end{equation*}
is defined for the augmented system \eqref{eq:augmented-system}, with $\Ap =
\frac{\de}{\de z}\tilde f$, $\Bp = \frac{\de}{\de u}\tilde f$ and $\Wp_j =
\blkdiag(W_j,\vec{0})$. The approximation to the overall cost is then $\Jn^*_k
+ J^d_k$, which is used in the subsequent optimization procedure.

\subsection{Optimization of the Current Control Input Gives Approximate
Dual Control}
\label{sec:optim-curr-contr}

The last step \ding{174} amounts to the outer loop of the overall algorithm. A
gradient-free black-box optimization algorithm is used to find the minimum of
the dual cost function. In every step, this algorithm proposes a control input
$u_k$ for which the dual cost is evaluated.

Depending on $u_k$, approximate filtering is carried out to the horizon.
The perturbation control is plugged into
Eq.~\eqref{eq:optimal-perturbation-cost}
to give an analytic, recursive definition for $\tilde K_j$, and an approximation
for the dual cost $J_k^d$, as a function of
the current control input $u_k$.

Nonlinear optimization---through repetitions of steps \ding{172} and \ding{173}
for proposed locations $u_k$---then yields an approximation to the optimal dual
control $u_k^*$. Conceptually the simplest part of the algorithm, this outer
loop dominates computational cost, because for every location $u_k$ the whole
machinery of \ding{172} and \ding{173} has to be evaluated.


\section{Extension to Contemporary Machine Learning Models}
\label{sec:extension-nonlinear-systems}

The preceding section reviewed the treatment of dual control in linear
dynamical systems from \cite{tse1973actively}. In this section, we extend the
approach to inference on, and dual control of, the dynamics of \emph{non}linear
dynamical systems. This extension is guided by the desire to use a number of
popular, standard regression frameworks in machine learning: Parametric general
least-squares regression, nonparametric Gaussian process regression, and
feedforward neural networks (including the base case of logistic regression).

\subsection{Parametric Nonlinear Systems}
\label{sec:param-nonl-syst}

We begin with the generalized linear model mentioned in
Eq.~\eqref{eq:feature-linear-system}. The nonlinear features $\phi$ can in
principle be any function (popular choices include sines and cosines, radial
basis functions, sigmoids, polynomials and others), with the caveat that their
structure crucially influences the properties of the model. From a modeling
perspective, this approach is quite standard for machine learning. However, the
dynamical learning setting requires a few adaptations: First, to allow the
modeling of higher-order dynamical systems, the original states must to be
included. This gives features of the form $\phi(x)\T = \begin{pmatrix} x\T &
\varphi(x)\T\end{pmatrix}$, consisting of the linear representation, augmented
by general features $\varphi$.

The next challenge is that the optimal control for nonlinear dynamical systems
cannot be optimized in closed form using dynamic programming, not even for the
deterministic nominal system. Instead, we find the nominal reference trajectory
using nonlinear model predictive control \citep{allgower1999nonlinear,
diehl2009nonlinear}. In our case, we begin with dynamic programming on a
locally linearized system, then optimize nonlinearly with a numerical method
across the trajectory. This adds computational cost, and requires some care to
achieve stable optimization performance for specific system setups.

Filtering from observations is also more involved in the case of nonlinear
dynamics. In the experiments reported below, we stayed within the extended
Kalman filtering framework to retain Gaussian beliefs over the states and
parameters. Extensions of this approach to more elaborate filtering methods are
an interesting direction for future work. This includes relatively standard
options like unscented Kalman filtering \citep{Uhlmann1995Dynamic}, but also
more recent developments in machine learning and probabilistic control, such as
analytic moment propagation where the features $\varphi$ allow this
\citep[e.g.,][]{Deisenroth2011c}.

The final problem is the generalization of the derivations from the preceding
sections to the nonlinear dynamics. We take a relatively simplistic approach,
which nevertheless turns out to work well. A linearization gives locally
linear dynamics whose structure closely matches Eq.~(\ref{eq:augmented-system}):
\begin{equation*}
  \begin{split}
    z_{k+1} &=
    \begin{pmatrix}
      \bar{x}_{k+1} + \Delta x_{k+1} \\\bar{\theta}_{k+1} + \Delta\theta_{k+1}
    \end{pmatrix} =
    \begin{pmatrix}
      A(\theta_k) & 0 \\ 0 & I
    \end{pmatrix}
    \begin{pmatrix}
      \phi(\bar{x}_k+\Delta x_k)\\ \bar{\theta}_k + \Delta\theta_k
    \end{pmatrix}
    +
    \begin{pmatrix}
      B(\theta_k) \\ 0
    \end{pmatrix}u_k +
    \begin{pmatrix}
      \xi_k \\ 0
    \end{pmatrix}\\
    &\approx
    \begin{pmatrix}
      A(\bar\theta_k) & 0 \\ 0 & I
    \end{pmatrix}
    \begin{pmatrix}
      \phi(\bar x_k)\\ \bar{\theta}_k
    \end{pmatrix} +
    \begin{pmatrix}
      B(\bar\t_k) \\ 0
    \end{pmatrix}u_k +
    \begin{pmatrix}
      \xi_k \\ 0
    \end{pmatrix}
    \\
    &\qqqq+
    \begin{pmatrix}
      A(\bar{\theta}_k)\frac{\de}{\de x_k}\phi(\bar x_k) &
      \frac{\de}{\de \t_k}\left( A(\bar \t_k) \phi(\bar x_k)
      + B(\bar\t_k)u_k\right)\\
      0 & I
    \end{pmatrix}
    \begin{pmatrix}
      \Delta x_k \\ \Delta \theta_k
    \end{pmatrix}.
  \end{split}
\end{equation*}
This essentially amounts to extended Kalman filtering on the augmented
state. Using this linearization, the approximation described in
Sec.~\ref{sec:appr-dual-contr} can be applied analogously.

\subsection{Nonparametric Gaussian Process Dynamics Models}
\label{sec:nonp-gauss-proc}

The above treatment of parametric linear models makes it comparably easy to
extend the description from finitely many feature functions to an
infinite-dimensional feature space defining a Gaussian process (GP) dynamics
model: Assume that the true dynamics function $f$ is a draw from a Gaussian
process prior $p(f) = \GP(f;m,\bar\kappa)$ with prior mean function $m:\Re^n
\to \Re^n$, and prior covariance function (kernel) $\bar{\kappa}:\Re^n\times
\Re^n\to \Re^n\times \Re^n$. This is using the widely used notion of
``multi-output regression'' \cite[][\textsection~9.1]{RasmussenWilliams}, \ie
formulating the covariance as
\begin{equation*}
  \label{eq:9}
  \cov(f_i(x),f_j(x')) = \bar\kappa_{ij}(x,x').
\end{equation*}
To simplify the treatment, we will assume that the covariance factors between
inputs and outputs, \ie $\bar\kappa_{ij}(x,x') = V_{ij}\kappa(x,x')$ with a
univariate kernel $\kappa:\Re^n\times \Re^n \to \Re$ and a positive
semi-definite matrix $V\in\Re^{n\times n}$ of output covariances. By Mercer's
theorem \citep[e.g.,][]{koenig86:_eigen_distr_compac_operat,RasmussenWilliams},
the kernel can be decomposed into a converging series over eigenfunctions
$\phi(x)$, as
\begin{equation}
  \label{eq:10}
  \kappa(x,x') = \sum_{\ell=1} ^\infty \lambda_\ell \phi _\ell(x) \phi_\ell ^*
(x'),
\end{equation}
where $\phi_\ell:\Re^n\to\Re$ are functions that are orthonormal relative to
some measure $\mu$ over $\Re^n$ (the precise choice of which is irrelevant for
the time being), with the property
\begin{equation*}
  \label{eq:11}
  \int \kappa(x,x') \phi_\ell(x') \, d\mu(x') = \lambda_\ell \phi_\ell(x).
\end{equation*}
Precisely in this sense, Gaussian process regression can be written as
``infinite-dimensional'' Bayesian linear regression. We will use the suggestive,
and somewhat abusive notation $f_k(x_k) = L\Omega_k\Phi(x_k)$ for this
generative model, defined as
\begin{equation}
  \label{eq:12}
  f^i _k(x_k) = \sum_{j=1} ^T L_{ij}  \sum_{\ell=1} ^\infty
\Omega_k ^{j\ell} \phi_i(x_k)
\end{equation}
where $L$ is a matrix satisfying $LL\T=V$ (\eg the Cholesky decomposition), and
the elements of $\Omega$ are draws from the ``white'' Gaussian process
$\Omega_k ^{j\ell} \sim \N(0,\lambda_\ell)$. Because of Mercer's theorem above,
Eq.~(\ref{eq:12}) exists in $\mu^2$ expectation, and is well-defined in this
sense. This notation allows writing Eq.~(\ref{eq:13}) as a nonparametric prior
with mean $\hat{\theta}_0$ and covariance $\Sigma_0 ^{\theta\theta} = V\otimes
(\Phi\Lambda\Phi \T)$ where $\Lambda$ is an infinite diagonal matrix with
diagonal elements $\Lambda_{\ell\ell} = \lambda_\ell$ (the matrix multiplication
$\Phi\Lambda\Phi\T$ is here defined as in Eq.~\eqref{eq:12}).

Using this notation, a tedious but straightforward linear algebra derivation
(see Appendix~\ref{sec:appendix-a}) shows that the posterior over
$z\T=\begin{pmatrix}x\T & \theta\T\end{pmatrix}$ after a number $k$ of
EKF-linearized Gaussian observations is a tractable Gaussian process, for which
the Gram matrix
\begin{equation*}
  \G = \P + \Q + \K + \F\inv\R\F^{-\intercal}
\end{equation*}
consists of the parts
\begin{equation*}
    \P = \begin{bmatrix} P_0 & 0 \\ 0 & 0 \end{bmatrix} \qq
    \Q = \left(Q\otimes I\right) \qq
    \K = \kappa(\vec{y}_{1:m},\vec{y}_{1:m}) \qq
    \R = \left(R\otimes I\right)
\end{equation*}
of appropriate size, depending on the current time $k$.
The multi-step state transition matrix
\begin{equation*}
  \F = \begin{bmatrix}
    I & 0 & 0 &  \cdots & 0 \\
    A_1 & I & 0 &  \cdots & 0 \\
    A_2 A_1 & A_2 & I &  \cdots & 0 \\
    \vdots & & & \ddots & \\
    A_m \cdots A_1 & A_m \cdots A_2 & \cdots & & I \\
    \end{bmatrix}
    \qq \text{with} \qq
    \F\inv =
    \begin{bmatrix}
    I & 0 & 0  & \cdots & 0 \\
    -A_1 & I & 0  & \cdots & 0 \\
    0 & -A_2 & I  & \cdots & 0 \\
    \vdots & \ddots & \ddots & \ddots & \\
    0 & \cdots &  0 &  -A_m & I \\
    \end{bmatrix}
\end{equation*}
is needed to account for the effect of the measurement noise $R$ over time.
The $A$-matrices are the Jacobians $\left.\nabla_x
f(x)\right|_{x_k}$.

The posterior mean now evaluates to
\begin{multline}
  \label{eq:np-mean-posterior}
  \begin{bmatrix}
    \hat{x}_k\\
    \hat{\theta}_k
  \end{bmatrix} =
  \begin{bmatrix}
    \hat x_{k-1} \\
    0
  \end{bmatrix}
  +
  \begin{bmatrix}
    \Phi(\hat x_{k-1})\Lambda\Phi(\vec{y}_{1:k-1})\T \\
    \Lambda\Phi(\vec{y}_{1:k-1})\T
  \end{bmatrix}
  \G\inv
  \begin{bmatrix}
    \Phi(\vec{y}_{1:k-1})\Lambda\Phi(\hat x_{k-1})\T &
    \Phi(\vec{y}_{1:k-1})\Lambda
  \end{bmatrix}
\end{multline}
and the posterior covariance is comprised of
\begin{subequations}
\begin{align}
  \bar \Sigma_k ^{xx} &= A_k \Sigma_k^{xx} A_k\T + Q + \Phi_k \Sigma^{\t x}_k
  A_k\T
  + A_k \Sigma^{x\t}_k \Phi_k\T + \Phi_k \Sigma^{\t\t}\Phi_k\T \\
  \Sigma_k^{xx} &= \bar\Sigma_{k-1} ^{xx} - \bar\Sigma_{k-1} ^{xx}
  \left[ \bar\Sigma_{k-1} ^{xx} + R \right]\inv \bar\Sigma_{k-1} ^{xx}\\
  \Sigma_k ^{x\theta} &= \F_{k,:}\Phi(\vec{y})\Lambda
   - \F_{k,:} \left[ \P + \K + \Q \right] \G\inv \Phi(\vec{y}) \Lambda
   \label{eq:np-cross-posterior}\\
  \Sigma_k ^{\theta x} &= (\Sigma_k ^{x\theta})\T\\
  \Sigma_k ^{\t\t} &= \Lambda
  - \Lambda \Phi(\vec{y})\T \G\inv \Phi(\vec{y}) \Lambda.
\label{eq:np-theta-posterior}
\end{align}
\end{subequations}

This formulation, together with the expositions in the preceding
sections, defines a nonparametric dual control algorithm for Gaussian
process priors. It is important to stress that this posterior is indeed
``tractable'' in so far as it depends only on a Gram matrix of size $nT\times
nT$, and the posterior over any $f(x)$ can be computed in time
$\mathcal{O}((nT)^3)$,
despite the infinite-dimensional state space.

\subsubsection{An Approximation of Constant Cost}
\label{sec:an-approximation-of-constant-cost}

\begin{figure}
  \setlength\figureheight{6cm}
  \setlength\figurewidth{0.9\columnwidth}
  \scriptsize
   \beginpgfgraphicnamed{kernel_comparison-external}%
   \input{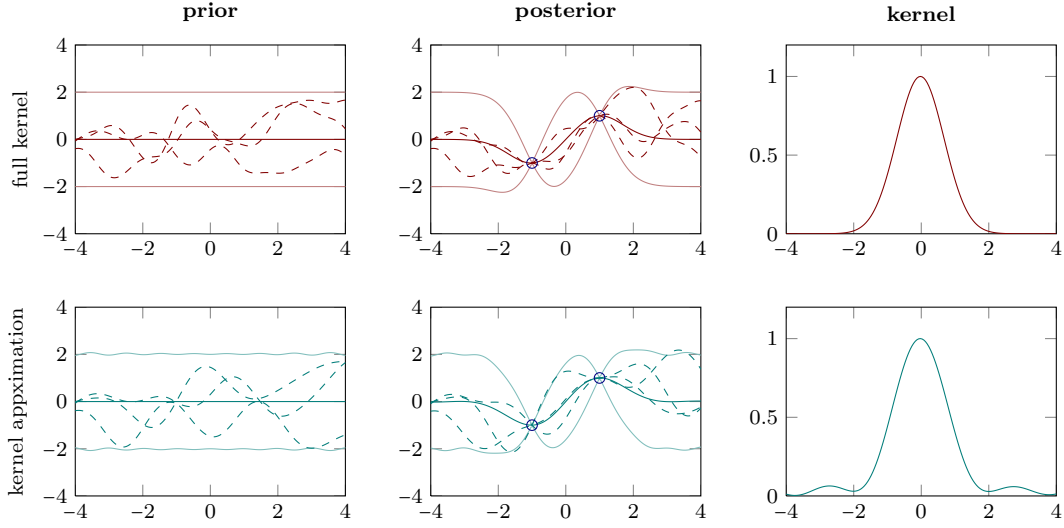}%
   \endpgfgraphicnamed%
 
  \caption{Prior (left), posterior (middle) and kernel function (right) of both
  the full kernel function (top row) and the approximate kernel (bottom row).
  The thick lines represent the mean and the thin lines show two standard
  deviations. The dashed lines are samples from the shown distributions.}
  \label{fig:kernel-approximation}
\end{figure}

In practical control applications, continuously rising inference
cost is rarely acceptable. It is thus necessary to project the GP belief
onto a finite representation, replacing the infinite sum in
Eq.~(\ref{eq:10}) with a finite one, to bound the computational cost
of the matrix inversion in Eqs.~\eqref{eq:np-mean-posterior},
\eqref{eq:np-cross-posterior} and~\eqref{eq:np-theta-posterior}. We do so by
projecting into a pre-defined finite basis of functions drawn from the
eigen-spectrum of the kernel with respect to the Lebesgue measure. This approach
has been recently popular elsewhere in regression \citep{RahimiRecht2008Random}.
For readers unaware of this line of work, here is a short, self-contained
introduction:

By Bochner's theorem \citep[e.g.,][]{stein1999interpolation, RasmussenWilliams},
the covariance function $k(r)$ (with $r=|x-x'|$) of a stationary $\mu^2$
continuous random process can be represented as the Fourier transform of a
positive finite measure and, if that measure has a density $S(s)$, as the
Fourier dual of $S$:
\begin{equation*}
  \kappa(r) = \int S(s)e^{2\pi i s r} ds,
\end{equation*}
This means that the eigenfunctions of the kernel are trigonometric functions,
and stationary covariance functions, like the commonly used square exponential
kernel
\begin{equation*}
  \kappa_\SE(x,x') = \exp\left(-\frac{(x-x')^2}{2\lambda^2}\right),
\end{equation*}
can be approximated by sine and cosine basis functions as
\begin{equation*}
  \kappa(x,x') \approx \tilde \kappa(x,x') =
\sqrt{\frac{2}{F}}\sum_{i=1}^{\nicefrac{F}{2}}
    \sin(\omega_{2i-1} |x-x'|) + \cos(\omega_{2i} |x-x'|),
\end{equation*}
where the frequencies $\omega_i$ of the feature functions is sampled from the
power spectrum of the process. An example of such kernel approximation is shown
in Fig.~\ref{fig:kernel-approximation}. With increasing number of features, the
approximation can be chosen as closely to the true covariance function as
needed, while keeping the number of features in a range that is still feasible
within the time constraints of the control algorithm.

\subsection{Dual Control of Feedforward Neural Networks}
\label{sec:dual-control-of-feedforward-neural-networks}

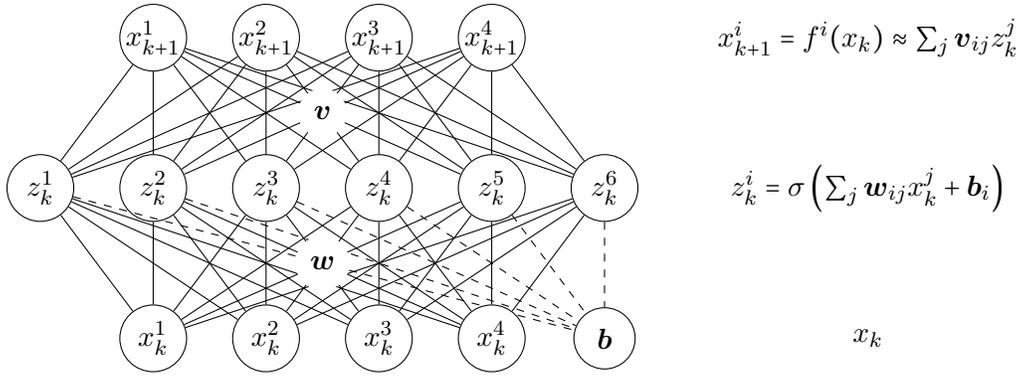
\begin{figure}
  \normalsize
  \centering
  \begin{tikzpicture}

    \foreach \x/\i in {2.5/1,4/2,5.5/3,7/4}
    {\node[var] at (\x,0) (f\i) {$x_{k+1} ^\i$};}

    \foreach \x/\i in {1/1,2.5/2,4/3,5.5/4,7/5,8.5/6}
    {\node[var] at (\x,-2) (h\i) {$z^\i _k$};

    \foreach \y in {1,2,3,4}
    {\draw (h\i) -- (f\y); }
  };

    \node[var] at (8.5,-4) (b) {$\vec{b}$};
    \foreach \i in {1,2,3,4,5,6}
    \draw[dashed] (h\i) -- (b);

    \foreach \x/\i in {2.5/1,4/2,5.5/3,7/4}
    {\node[var] at (\x,-4) (x\i) {$x_{k} ^\i$};
      \foreach \y in {1,2,3,4,5,6}
      {\draw (x\i) -- (h\y);}
    }

    \node at (12,-4) {$x_k$};
    \node[shape=circle,fill=white] at (4.75,-3) {$\vec{w}$};
    \node at (12,-2) {$z_k ^i = \sigma\left( \sum_{j} \vec{w}_{ij} x ^j _k
+
    \vec{b}_i \right)$};
    \node[shape=circle,fill=white] at (4.75,-1) {$\vec{v}$};
    \node at (12,0) {$x_{k+1} ^i = f^i(x_k)\approx \sum_j \vec{v}_{ij} z_k ^j$};
  \end{tikzpicture}
  \caption{A two-layer feedforward neural network. Sketch to illustrate the
structure of Eq.~\eqref{eq:8}.}
  \label{fig:NN}
\end{figure}

Another extension of the parametric linear models of
Section~\ref{sec:param-nonl-syst} is to allow for a nonlinear parametrization
of the dynamics function:
\begin{equation*}
  \label{eq:7}
  f(x;\theta) = \sum_i^F \theta^\text{lin} _i \phi_i(x;\theta_i ^\text{nonlin}).
\end{equation*}
A particularly interesting example of this structure are multilayer perceptrons.
Consider a two-layer network with logistic link function
\begin{equation}
  \label{eq:8}
  f(x) = \sum_i \vec{v}_i \sigma(\vec{w}_ix + \vec{b}_i),
\end{equation}
where $\vec{v}$ are the weights from the latent to the output layer, $\vec{w}$
are the weights from input to hidden units, and $\vec{b}$ are the biases of
the hidden units (see Fig.~\ref{fig:NN}).

Neural networks are used in control quite regularly, see \eg
\cite{NguyenWidrow1990Neural}. Instead of using backpropagation and stochastic
gradient descent as in most applications of neural networks
\citep{rumelhart1986learning, robbins1951stochastic}, the EKF inference
procedure can be used to train the weights as well
\citep{SinghalWu1989Training}. This is possible because the EKF linearization
can also be applied for the nonlinear link function, \eg the logistic function.
Speaking in terms of feature functions, not only the weight of each feature but
also the shape (steepness) can be inferred. A limiting factor for this inference
naturally is the number of data points: the more features and parameters are
introduced, the more data points are necessary to learn.

Using the state augmentation $z\T = (x\T \; v\T \; w\T)$, and linearizing \wrt
all parameters in each step, the EKF inference on the neural network parameters
allows us not only to apply relatively cheap inference on them, but also to use
the dual control framework to plan control signals, accounting for the effect
of future observations and the subsequent change in the belief. This means the
adaptive dual controller described in Sec.~\ref{sec:appr-dual-contr} can
identify those parts of the neural net that are relevant for applying optimal
control to the problem at hand. In Sec.~\ref{sec:exp-information-importance},
we show an experiment with these properties.

\section{Experiments}
\label{sec:experiments}

A series of experiments on single-episode tasks with continuous state space
highlights qualitative differences between the adaptive dual (AD) controller
and three other controllers: An oracle controller with access to the true
parameters, which provides an unattainable lower bound (LB) on the achievable
performance, a certainty equivalent (CE) controller as described
in Sec.~\ref{sec:cert-equiv-contr}, and a controller minimizing the sum of CE
cost and the Bayesian exploration bonus (BEB)
\citep[see][]{kolter2009near}:
\begin{equation*}
  \ell_\BEB =
  \tau
\left[\operatorname{sqrt}\left(\diag\left(\Sigma^{\tt}\right)\right)\right]\T
\left[\operatorname{sqrt}\left(\diag\left(\Sigma^{\tt}\right)\right)\right]
\end{equation*}
($\tau$ is a scalar exploration weight).
The additional cost term $\ell_\BEB$ is evaluated for the predicted parameter
covariance where the prediction time is chosen according to the order of the
system such that the effect of the current control signal shows up in the
belief over the parameters. This type of controller is sometimes also called
dual control, while being referred to as \emph{explicit dual control}, where
the dual features are obtained by a modified cost function
\citep{FilatovU2000Survey}.

Every experiment was repeated $50$ times with different random seeds, which
were shared across controllers for comparability. All systems presented
below are very simple setups. Their primary point is to show qualitative
differences of the controllers' behavior. The experiments were done with
different approximations from the preceding section to show experimental
feasibility for each of them.

The feature set used for a specific application is part of the prior
assumptions for that application. Large uncertainty requires flexible models
(which take longer to converge, and require more exploration). Feature
selection is important, but since it is independent of the dual
control framework itself and a broad topic on its own, it is beyond the scope
of this paper. In the following experiments, different feature sets are used
both as examples for the flexibility of the framework, but also to model
different structural knowledge about the problems at hand.

\subsection{On a Simple Scalar System, AD Control Matches Exact Dual Control
Well}
\label{sec:exp-simple-scalar}

For the noise-free linear system of Sec.~\ref{sec:toy-problem}, ($a = 1$
(known), $b = 2$, $p(b)=\N(b; 1, 10)$, $Q = 10^{-1}$, $R = 0$, $W = 1$,
$\L = 1$, $T=2$), Fig.~\ref{fig:sampling_uncertain_b}, right, compares the cost
functions of the various controllers and the approximately exact sampling
solution (which is only available for this very simple setup). All cost
functions are shifted by an irrelevant constant. The CE cost is quadratic and
indifferent about zero. The BEB ($\tau = 0.1$) gives additional structure near
zero that encourages learning. While qualitatively similar to the dual cost,
its global minimum is almost at the same location as that of CE. The dual
control approximates the sampling solution much closer.

\subsection{Faced with Time-Varying State Cost, AD Control Holds Off
  Exploration Until Suitable}
\label{sec:exp-pendulum}

\begin{figure}[ht]
  \setlength\figureheight{8cm}
  \setlength\figurewidth{0.96\columnwidth}
  \scriptsize
   \beginpgfgraphicnamed{time_relevance_mod-external}%
   \input{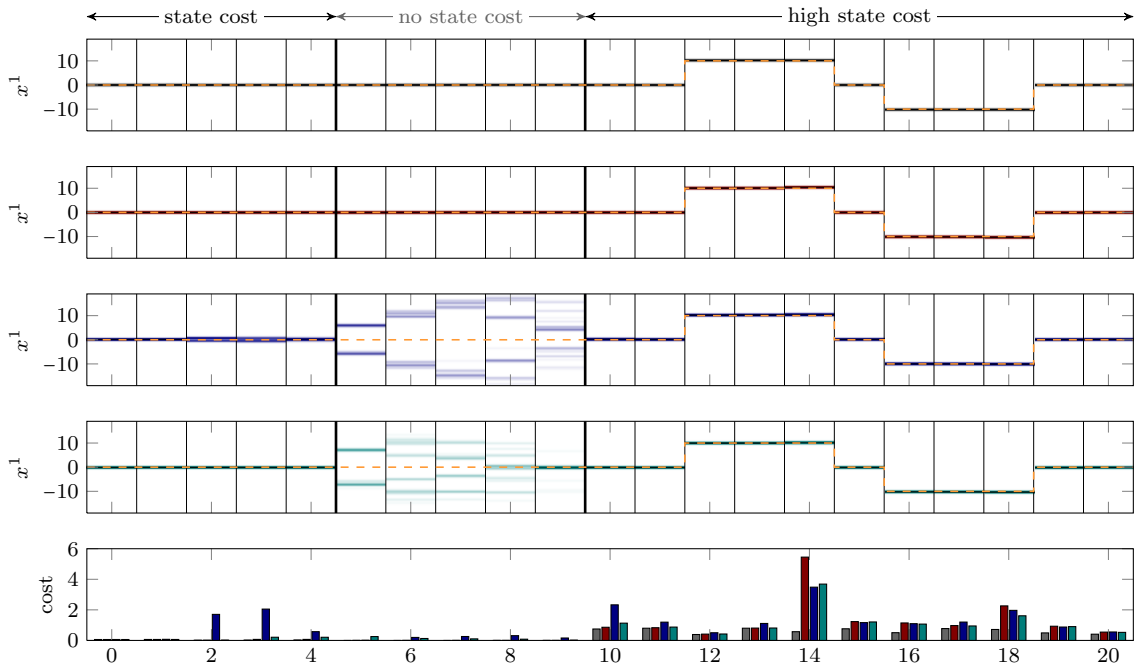}%
   \endpgfgraphicnamed%
 
  \caption{{\bfseries Top four:} Density estimate for 50 trajectories (first
state). From top to bottom: optimal oracle control (gray), certainty equivalent
control (red), CE with Bayesian exploration bonus (blue), approximate dual
control (green). Reference trajectory in dashed orange. {\bfseries Bottom:} The
mean cost per time step is shown in the bottom plot, with colors matching
the controllers noted above.
  }\label{fig:controller-comparison-pendulum}
\end{figure}

A cart on a rail is a simple example for a dynamical system. Combined with
a nonlinearly varying slope, a simple but nonlinear system can be constructed.
The dynamics, prior beliefs, and true values for the parameters are chosen to be
\begin{equation*}
  \label{eq:pendulum-system}
  x_{k+1} = \begin{bmatrix} 1 & 0.4 \\ 0 & 1 \end{bmatrix} x_k
  + \begin{bmatrix} 0 & 0 \\ \t^1 & \t^2 \end{bmatrix}
  \begin{bmatrix} \varphi^1(x^1_k) \\ \varphi^2(x^1_k) \end{bmatrix}
  + \begin{bmatrix} 0 \\ 1 \end{bmatrix} u_k
  \q
  \t \sim \N\left(\begin{bmatrix} 0 \\ 0 \end{bmatrix},
    \begin{bmatrix}1 & 0 \\ 0 & 1  \end{bmatrix}\right)
  \q
  \t_\text{true} = \begin{bmatrix} 0.8 \\ 0.4 \end{bmatrix},
\end{equation*}
where superscripts denote vector elements. The nonlinear functions $\varphi$
are shifted logistic functions of the form
\begin{equation}
  \label{eq:logistic-nonlinearity}
  \varphi^1(x) = -\frac{1}{1+e^{(x+5)}} \qq
  \varphi^2(x) = \frac{1}{1+e^{-(x-5)}},
\end{equation}
and disturbance/noise is chosen to be $R = Q = 10^{-2} I$. We use this setup as
a testbed for a time-structured exploration problem. The actual system and its
dynamics are relatively irrelevant here, as we will focus on a complication
caused by the cost function: The reference to be tracked is
\begin{equation*}
  \label{eq:exp2-trajectory}
  \vec{r}_{0:11} = \begin{bmatrix} 0 \\ 0 \end{bmatrix}
  \qquad
  \vec{r}_{12:14} = \begin{bmatrix} 10 \\ 0 \end{bmatrix}
  \qquad
  \vec{r}_{15} = \begin{bmatrix} 0 \\ 0 \end{bmatrix}
  \qquad
  \vec{r}_{16:18} = \begin{bmatrix} -10 \\ 0 \end{bmatrix}
  \qquad
  \vec{r}_{19:20} = \begin{bmatrix} 0 \\ 0 \end{bmatrix};
\end{equation*}
it is also shown in each plot of Fig.~\ref{fig:controller-comparison-pendulum}
as dashed orange line. The state weighting is time-dependent
\begin{equation*}
  \vec{W}_{0:5} = \begin{bmatrix} 10 & 0 \\ 0 & 0 \end{bmatrix}
  \qquad
  \vec{W}_{5:10} = \begin{bmatrix} 0 & 0 \\ 0 & 0 \end{bmatrix}
  \qquad
  \vec{W}_{11:20} = \begin{bmatrix} 100 & 0 \\ 0 & 0 \end{bmatrix},
\end{equation*}
and control cost is relatively low: $\L = 10^{-3}$. The task, thus, is to first
keep the cart fixed in the starting position to high precision, for the
first 4 time steps. This is followed by a ``loose'' period between time steps
$5$ and $10$. Then, the cart has to be moved to one side, back to the center, to
the other side, and back again, all at high cost. A good exploration strategy in
this setting is to act cautiously for the first 5 time steps, then aggressively
explore in the ``loose'' phase, to finally be able to control the motion with
high precision.

The inference model is a GP with approximated SE kernel, as described in
Sec.~\ref{sec:an-approximation-of-constant-cost}. We use $30$ alternating sine
and cosine features that are distributed according to the power spectrum of the
full SE kernel. Since the true nonlinearity of
Eq.~\eqref{eq:logistic-nonlinearity} is not of this form, the approximation is
out of model and the lower bound controller only represents a perfectly
learned, but still not exact, model.

Fig.~\ref{fig:controller-comparison-pendulum} shows a density estimated from 50
state trajectories for the four different controllers. The lower bound
controller (top) controls precisely at times of high cost, and does nothing for
times with zero cost, controlling perfectly up to the measurement and state
disturbances. The certainty equivalent controller (second from top) never
explores actively, it only learns ``accidentally'' from observations arising
during the run. Since the initial trajectory requires little action, it is left
with a bad model when the reference starts to move at time step $12$. The
exploration bonus controller (second from bottom) continuously explores,
because it has no way of knowing about the ``loose'' phase ahead. Of course,
this strategy incurs a higher cost initially. The dual controller (bottom)
efficiently holds off exploration until it reaches the ``loose'' phase, where
it explores aggressively.

\subsection{AD Control Distinguishes Necessary and Unnecessary Parameter
Exploration}
\label{sec:exp-information-importance}
The system including nonlinearities for this experiment is the same as before,
although with noise parameters $R = Q = 10^{-3}I$. The reference
trajectory and state weighting are much simpler, though:
\begin{equation*}
  \label{eq:exp3-trajectory}
  \vec{r}_{0:11} = \begin{bmatrix} 0 \\ 0 \end{bmatrix}
  \qquad
  \vec{r}_{12:18} = \begin{bmatrix} 10 \\ 0 \end{bmatrix}
  \qquad
  \vec{r}_{18:20} = \begin{bmatrix} 0 \\ 0 \end{bmatrix},
\end{equation*}
with the time-dependent weighting
\begin{equation*}
  \vec{W}_{0:10} = \begin{bmatrix} 0 & 0 \\ 0 & 0 \end{bmatrix}
  \qquad
  \vec{W}_{11:20} = \begin{bmatrix} 10 & 0 \\ 0 & 0 \end{bmatrix},
\end{equation*}
allowing for identification in the beginning, while penalizing deviations of
the first state in later time steps.

Important to note here is that the reference trajectory only passes areas of
the state space where $\varphi^1$ is strong, and $\varphi^2$ is negligible.
Good exploration thus will ignore $\theta^2$, but this can only be found
through reasoning about future trajectories.

In this experiment, the learned model is of the neural network form described
in Sec.~\ref{sec:dual-control-of-feedforward-neural-networks}. We use $4$
logistic features (see Eq.~\eqref{eq:8}) with two free parameters each ($w_i$
and $v_i$) and equally spaced $b_i$ between $-5$ and $5$, the locations of the
true nonlinear features. This means it is possible to learn the perfect model in
this case.

\begin{figure}[ht]
  \setlength\figureheight{8cm}
  \setlength\figurewidth{0.95\columnwidth}
  \scriptsize
   \beginpgfgraphicnamed{position_relevance_mod-external}%
   \input{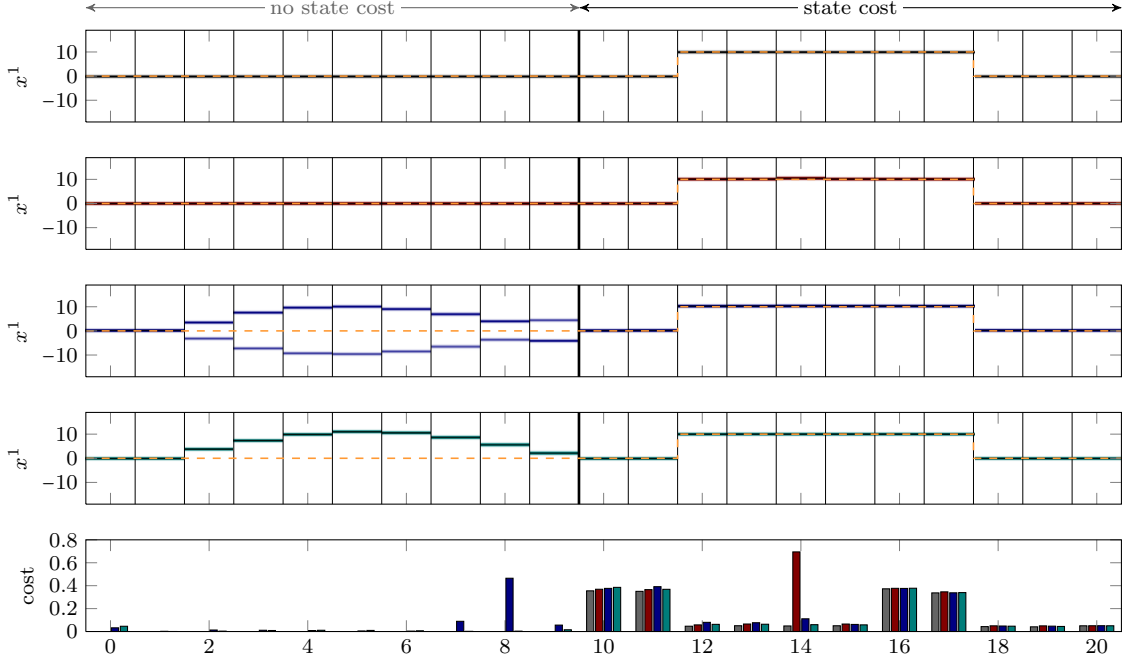}%
   \endpgfgraphicnamed%
 
  \caption{{\bfseries Top four:} Density estimate for 50 trajectories (second
state). From top to bottom: optimal oracle control (gray), certainty equivalent
control (red), CE with Bayesian exploration bonus (blue), approximate dual
control (green). Reference trajectory in dashed orange. {\bfseries Bottom:} The
mean cost per time step is shown in the bottom plot, with colors matching
the controllers noted above.
  }\label{fig:position-relevance}
\end{figure}

Fig.~\ref{fig:position-relevance} shows a density estimated from 50 state
trajectories for the four different controllers. Because of symmetry in the
cost function and feature functions, BEB (with $\tau=1$) cannot ``decide''
between the relevant $\theta^1$ and the irrelevant $\theta^2$, choosing the
exploration direction stochastically. It thus sometimes reduces the uncertainty
on $\t^2$, which does not help the subsequent control. The AD controller ignores
$\t^2$ completely and only identifies $\t^1$ in early phases, leading to good
control performance.

\subsection{AD Control Maintains Only Useful Knowledge}
\label{sec:exp-knowledge-maintenance}
The last experiment is again similar to Sec.~\ref{sec:exp-pendulum}, but uses a
different set of nonlinear functions, namely shifted Gaussian functions
(\aka radial basis functions)
\begin{equation*}
  \varphi^1(x) = e^{-\frac{(x-2)^2}{2}}
  \qq
  \varphi^2(x) = e^{-\frac{(x+2)^2}{2}}
  \qq
  \t_\text{true} = \begin{bmatrix} 1.0 \\ 0.8 \end{bmatrix}.
\end{equation*}
For this experiment, the model is learned with parametric linear regression,
according to Sec.~\ref{sec:param-nonl-syst}. The fundamental difference to the
other experimental setups is that the model now assumes \emph{parameter drift}.
This results in growing uncertainty for the parameters over time. (The true
parameters are kept constant for simplicity.)

The reference to be tracked passes through both nonlinear features but then
stays at one of them:
\begin{equation*}
  \label{eq:exp4-trajectory}
  \vec{r}_{0:6} = \begin{bmatrix} -5 \\ 0 \end{bmatrix}
  \qquad
  \vec{r}_{7} = \begin{bmatrix} -4 \\ 0 \end{bmatrix}
  \qq
  \vec{r}_{8} = \begin{bmatrix} -2 \\ 0 \end{bmatrix}
  \qq
  \vec{r}_{9} = \begin{bmatrix} 0 \\ 0 \end{bmatrix}
  \qquad
  \vec{r}_{10:20} = \begin{bmatrix} 2 \\ 0 \end{bmatrix}.
\end{equation*}
The cost structure is
\begin{equation*}
  \vec{W}_{0:5} = \begin{bmatrix} 0 & 0 \\ 0 & 0 \end{bmatrix}
  \qquad
  \vec{W}_{6:20} = \begin{bmatrix} 10 & 0 \\ 0 & 0 \end{bmatrix},
\end{equation*}
such that there is cost starting with the linear reference trajectory at time
instant $6$.

\begin{figure}
  \setlength\figureheight{6cm}
  \setlength\figurewidth{\columnwidth}
  \scriptsize
   \beginpgfgraphicnamed{information_maintenance-external}%
   \input{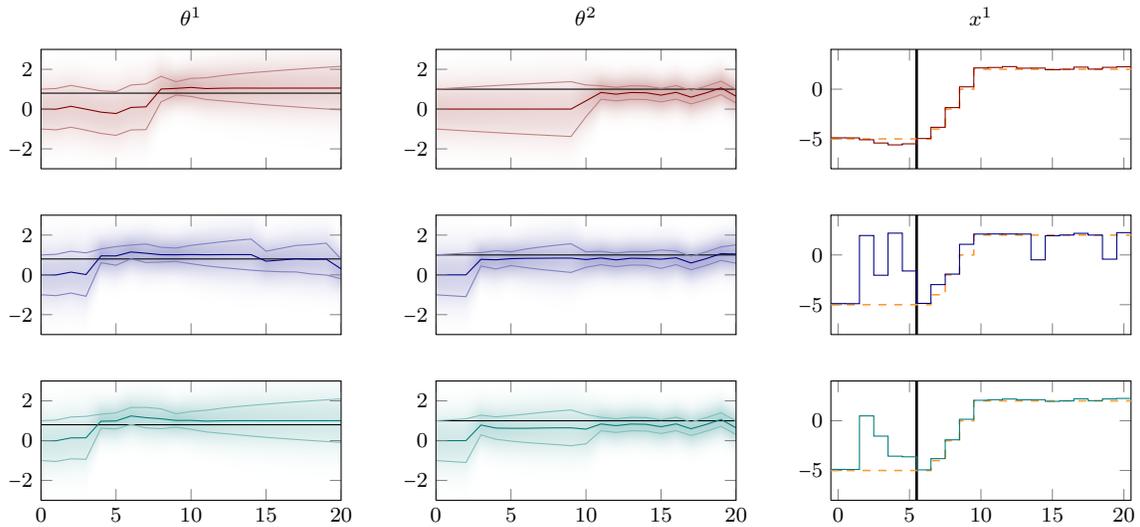}%
   \endpgfgraphicnamed%
 
  \caption{Parameter knowledge (left, middle) and state trajectory (right)
for different controllers. From top to bottom: certainty equivalent control
(red), CE with Bayesian exploration bonus (blue), approximate dual control
(green). The true parameters are the black lines.
  }\label{fig:information-maintenance}
\end{figure}

Fig.~\ref{fig:information-maintenance} shows the parameter belief and relevant
state of a single run of this experiment over time. It shows clearly that the in
the beginning necessary parameter $\t^1$ is learned early by BEB and the AD
controller, while CE learns only ``accidentally''. The BEB controller also
learns the second parameter in the beginning, even though the knowledge will be
lost over time. When the trajectory reaches the zone of the second parameter,
the BEB controller tries to lower the growing uncertainty every now and then
(visible by the drops in state $x^1$), incurring high cost. AD control
completely ignores the growing uncertainty on $\t^1$ after reaching the area of
$\t^2$, thus preventing unnecessary exploration.

\subsection{Quantitative Comparison}

The above experiments aim to emphasize qualitative strengths of AD control
over simpler approximations. It is desirable for a controllers to deal
with flexible models of many parameters, many of which will invariably
be superfluous. For reference, Table~\ref{tab:results} also shows
quantitative results: Averages and standard deviations of the cost,
from the 50 runs for each controller. The AD controller shows good performance
overall; interestingly, it also has low variance. CE and BEB were more
prone to instabilities.

\begin{table}
\begin{centering}\small
\begin{tabular}{@{} l | r r | r r | r r | r r @{}}
\toprule
& \multicolumn{2}{c|}{Exp.~\ref{sec:exp-simple-scalar}} &
\multicolumn{2}{c|}{Exp.~\ref{sec:exp-pendulum}} &
\multicolumn{2}{c|}{Exp.~\ref{sec:exp-information-importance}} &
\multicolumn{2}{c}{Exp.~\ref{sec:exp-knowledge-maintenance}}
\\       & mean&std  & mean&std  &  mean&std  &  mean&std \\
\midrule
Oracle   & 0.67&0.23 & 1.75&0.33 &  7.15&3.85 &  0.66&0.51 \\
CE       & 0.84&0.73 & 2.49&0.74 & 15.72&5.20 &  1.76&0.90 \\
CE-BEB   & 0.99&0.92 & 2.64&0.37 & 20.88&6.74 & 84.91&6.77 \\
AD       & 0.77&0.43 & 1.96&0.34 & 14.33&5.40 &  1.62&0.56 \\
\bottomrule
\end{tabular}\\
\caption{Average and standard deviation of costs in the experiments for 50
runs.}
\vskip -1.5mm
\end{centering}
\label{tab:results}
\end{table}
\section{Conclusion}
\label{sec:conclusion}

Bayesian reinforcement learning, or dual control, offers an elegant answer to
the exploration-exploitation trade-off, relative to prior probabilistic beliefs.
Its intricate, intractable structure requires approximations to balance another
kind of trade-off, between computation and performance. This work investigated
an old approximate framework from control, re-phrased it in the language of
reinforcement learning, and extended it to apply to contemporary inference
methods from machine learning, including approximate Gaussian process regression
and multi-layer networks. The result is a tractable approximation that captures
notions of structured exploration, like the value of waiting for future
exploration opportunities, and distinguishing relevant from irrelevant model
parameters.

The dual control framework, in its now clearer form, offers interesting
directions for research in reinforcement learning, including its combination
with recent new developments in learning and planning. Following this conceptual
work, the main challenge for further development is the still comparably high
(but tractable) numerical load of dual control, particularly in problems of
higher dimensionality.

\newpage

\appendix
\section{Nonparametric EKF Form}
\label{sec:appendix-a}

The standard Kalman filter (KF) can be found in many textbooks
\citep[e.g.,][]{sarkka2013bayesian} and therefore will not be restated here.
Starting from the standard equations, we derive a general multi-step formulation
of the classic KF with
\begin{equation*}
  p(z_k) = \N(z_k,m_k,P_k).
\end{equation*}
From there, state augmentation with an infinite-dimensional weight vector gives
the expected result.

\subsection{Derivation of the Multi-Step KF Formulation}
Assuming that the result of the KF and the Gaussian process framework should be
identical under certain circumstances, we wish to transform the KF to a
formulation with full Gram matrix. Therefore, the prediction and update step
have to be combined to
\begin{equation*}\label{eq:single-step}
  P_1 = \left(A_0 P_0 A_0\T + Q\right) - \left(A_0 P_0 A_0\T + Q\right)H\T
  S_1\inv H \left(A_0 P_0 A_0\T + Q\right)\T
\end{equation*}
\begin{equation*}
  S_1 = H\left(A_0 P_0 A_0\T + Q\right)H\T + R,
\end{equation*}
which is pretty straightforward. We're adopting a standard notation, where $P_k$
is the covariance at time step $k$, $A_k$ is the Jacobian, $Q$ is the drift and
$R$ is the measurement covariance.
The same can be done for the second time step, but it is beneficial
introducing a compact notation for the predictive covariance first
\begin{multline*}
\label{eq:compact-notation}
  \left(A_1  P_1  A_1\T + Q\right)\\
  =
  \left(A_1  \left[
  \left(A_0 P_0 A_0\T + Q\right) - \left(A_0 P_0 A_0\T + Q\right)H\T
    S_1\inv  H \left(A_0 P_0 A_0\T + Q\right)\T
  \right]  A_1\T + Q\right)\\
  =
  \underbrace{A_1 \left(A_0 P_0 A_0\T + Q\right) A_1\T + Q}_{\ce g_{11}}
   -
   \underbrace{A_1 \left(A_0 P_0 A_0\T + Q\right)}_{\ce g_{10}}H\T
  (\underbrace{S_1}_{\ce g_{00}})\inv
  H \underbrace{\left(A_0 P_0 A_0\T + Q\right) A_1\T}_{\ce g_{01}}
  \\
  =
  g_{11} - g_{10}H\T g_{00}\inv H g_{01}.
\end{multline*}
Using the compact notation and defining $S_2$ analogously to $S_1$, we can
write the two-step update as
\begin{equation*}
  P_2 = \left(g_{11} - g_{10}H\T S_1\inv H g_{01}\right)
  - \left(g_{11} - g_{10}H\T S_1\inv H g_{01}\right)
  H\T S_2\inv H \left(g_{11} - g_{10}H\T S_1\inv H g_{01}\right)
\end{equation*}
\begin{multline*}
  P_2 = g_{11} - g_{10}H\T S_1\inv H g_{01}
  - g_{11}H\T S_2\inv H g_{11}
  - g_{10}H\T S_1\inv H g_{01} H\T S_2\inv H g_{10}H\T S_1\inv H g_{01}\\
  + g_{11}H\T S_2\inv H g_{10}H\T S_1\inv H g_{01}
  + g_{10}H\T S_1\inv H g_{01} H\T S_2\inv H g_{11}
\end{multline*}
\begin{equation*}
  P_2 = g_{11} - \begin{bmatrix}g_{10}H\T & g_{11}H\T\end{bmatrix}
  \underbrace{
  \begin{bmatrix}
  S_1\inv + S_1\inv H g_{01} H\T S_2\inv H g_{10}H\T S_1\inv &
  -S_1\inv H g_{01} H\T S_2\inv \\
  -S_2\inv H g_{10} H\T S_1\inv&
  S_2\inv
  \end{bmatrix}
  }_{\ce G\inv}
  \begin{bmatrix}H g_{01} \\ H g_{11}\end{bmatrix}.
\end{equation*}
Application of Schur's lemma gives
\begin{equation*}
  G
  =
  \begin{bmatrix}
  H g_{00} H\T + R &
  H g_{01} H\T \\
  H g_{10} H\T &
  H g_{11} H\T + R
  \end{bmatrix}.
\end{equation*}

Assuming full state measurement ($H=I$) for compactness of notation, the
two-step update is
\begin{multline}
  P_2 = g_{11}
  - \begin{bmatrix}g_{10}\T & g_{11}\T\end{bmatrix}
    \begin{bmatrix}
  g_{00}+ R &
  g_{01}\\
  g_{10}&
  g_{11} + R
  \end{bmatrix}\inv
  \begin{bmatrix}g_{01} \\ g_{11}\end{bmatrix}\\
  =
  A_1 \left(A_0 P_0 A_0\T + Q\right) A_1\T + Q
  - \begin{bmatrix}A_1 \left(A_0 P_0 A_0\T + Q\right) &
    \left( A_1 \left(A_0 P_0 A_0\T + Q\right) A_1\T + Q \right)\end{bmatrix}
    \\
    \cdot
  \begin{bmatrix}
  \left(A_0 P_0 A_0\T + Q\right) + R &
  \left(A_0 P_0 A_0\T + Q\right) A_1\T \\
  A_1 \left(A_0 P_0 A_0\T + Q\right) &
  \left( A_1 \left(A_0 P_0 A_0\T + Q\right) A_1\T + Q \right) + R
  \end{bmatrix}\inv
  \begin{bmatrix}\left(A_0 P_0 A_0\T + Q\right) A_1\T \\
  \left( A_1 \left(A_0 P_0 A_0\T + Q\right) A_1\T + Q \right)\end{bmatrix},
  \label{eq:twostep-result}
\end{multline}
which already looks similar to GP inference. We can now generalize this two-step
result to the general form by building the Gram matrix according to
\begin{equation*}
  G = \F\P\F\T + \F\Q\F\T + \R,
\end{equation*}
where the individual parts are
\begin{equation*}
    \P = \begin{bmatrix} P_0 & 0 \\ 0 & 0 \end{bmatrix} \qq
    \Q = \left(Q\otimes I\right) \qq
    \R = \left(R\otimes I\right)
\end{equation*}
of appropriate size, depending on the current time $k$.
The multi-step state transition matrix
\begin{equation*}
  \F = \begin{bmatrix}
    I & 0 & 0 &  \cdots & 0 \\
    A_1 & I & 0 &  \cdots & 0 \\
    A_2 A_1 & A_2 & I &  \cdots & 0 \\
    \vdots & & & \ddots & \\
    A_m \cdots A_1 & A_m \cdots A_2 & \cdots & & I \\
    \end{bmatrix},
    \qq \text{with} \qq
    \F\inv =
    \begin{bmatrix}
    I & 0 & 0  & \cdots & 0 \\
    -A_1 & I & 0  & \cdots & 0 \\
    0 & -A_2 & I  & \cdots & 0 \\
    \vdots & \ddots & \ddots & \ddots & \\
    0 & \cdots &  0 &  -A_m & I \\
    \end{bmatrix},
\end{equation*}
is also needed so shift the initial covariance and drift covariances through
time. Put together, this results in
\begin{equation*}
  P_k = \F_{k,:}(\P + \Q)\F_{k,:}\T - \F_{k,:}(\P + \Q)\F (\F\P\F\T + \F\Q\F\T +
\R)\inv \F\T (\P + \Q) \F_{:,k}.
\end{equation*}
A more compact notation can be achieved by using $\F\inv$ to obtain
\begin{equation}
  \label{eq:cov-prediction}
  P_k = \F_{k,:}(\P + \Q)\F_{k,:}\T - \F_{k,:}(\P + \Q)(\P + \Q +
\F\inv\R\F^{-\intercal})\inv (\P + \Q) \F_{:,k}.
\end{equation}
Calculating the mean prediction is done analogously:
\begin{equation*}
  \label{eq:mean-prediction}
  m_k = \F_{k,0} m_0 + \F_{k,:}(\P + \Q)(\P + \Q +
\F\inv\R\F^{-\intercal})\inv \vec{y}.
\end{equation*}

\subsection{Augmenting the State}

Instead of tracking only the state covariance, in the GP setting also the
dynamics function has to be inferred.
The system equations of the nonlinear system are now
\begin{subequations}
\label{eq:system-equations-augmented}
\begin{align*}
  x_k &= f( x_{k-1}) + q_{k-1}  & q_{k-1}&\sim\N(0,Q)\\
  y_k &= H x_k + r_k  & r_k&\sim\N(0,R),
\end{align*}
\end{subequations}
where $f \sim \GP(0,k)$. The inference in this model can be done through the
EKF with augmented state. We adopt the weight-space view with $f =
\Phi\T w$ \citep[see][]{RasmussenWilliams} to augment the state with the
infinite-dimensional weight vector $w$:
\begin{equation*}
  z = \begin{pmatrix}x\\w\end{pmatrix} \qq
  \Sigma = \begin{pmatrix} P & \Sigma^{xw} \\ \Sigma^{wx} & \Sigma^{ww}
\end{pmatrix} \qq
  J_A = \begin{pmatrix} \frac{\de \bar f}{\de x} & \frac{\de \bar f}{\de w} \\
  0 & I \end{pmatrix}
  = \begin{pmatrix} A & \Phi \\
  0 & I \end{pmatrix},
\end{equation*}
where, in \eqref{eq:twostep-result}, the original $x$ is replaced by the
augmented $z$, $P$ by $\Sigma$ and $A$ by $J_A$.

Choosing $H=\begin{bmatrix}I & 0\end{bmatrix}$ so that $H$ recovers the
original states from the augmented state vector, we obtain, after calculations
similar to those above, a Gram matrix with additional terms including feature
functions and the prior on them:
\begin{equation*}
  G^\star = G + \begin{pmatrix}
  \Phi_0 \Sigma^{ww}_0 \Phi_0\T&
  \Phi_0 \Sigma^{ww}_0 \left(\Phi_0\T A_1\T + \Phi_1\T\right)\\
  \left(A_1 \Phi_0 + \Phi_1 \right) \Sigma^{ww}_0 \Phi_0\T &
  \left(A_1 \Phi_0 + \Phi_1 \right) \Sigma^{ww}_0 \left(\Phi_0\T A_1\T +
\Phi_1\T\right)
\end{pmatrix}.
\end{equation*}
At this point it is important to note that the infinite inner product $\Phi_k
\Sigma^{ww}_0 \Phi_k\T$ corresponds to an evaluation of the kernel
\begin{equation*}
  \Phi(x) \Sigma^{ww}_0 \Phi(x)\T = \sum \phi_i(x) \Sigma^{ww}_{0,ij} \phi_j(x')
=
\kappa(x,x').
\end{equation*}
This means we can write the Gram matrix as
\begin{equation*}
  G^\star = G + \begin{pmatrix}
  \kappa_{00} &
  \kappa_{00} A_1\T + \kappa_{01}\\
  A_1 \kappa_{00} + \kappa_{10} &
  A_1 \kappa_{00} A_1 \T + \kappa_{10} A_1\T + A_1
  \kappa_{01} + \kappa_{11}
\end{pmatrix},
\end{equation*}
where we have written $\kappa_{\cdot\cdot}$ for $\kappa(\cdot,\cdot)$ to save
space.
In total, the Gram matrix is then
\begin{equation*}
  G^{\star} = \F\P\F\T + \F\Q\F\T + \F\K\F\T + \R,
\end{equation*}
with $\K = \kappa(\vec{y}_{1:m},\vec{y}_{1:m})$. Since inference is more compact
and
numerically stable if we absorb $\F$ into the Gram matrix as in
Eq.~\eqref{eq:cov-prediction}, we define
\begin{equation*}
  \G = \P + \Q + \K + \F\inv\R\F^{-\intercal}.
\end{equation*}
Inference is done according to
\begin{subequations}
\begin{equation*}
  P_k = \F_{k,:}(\P + \Q + \K)\F_{k,:}\T - \F_{k,:}(\P + \Q + \K)\G\inv (\P +
\Q + \K) \F_{:,k}
\end{equation*}
\begin{equation*}
  \Sigma^{wx}_k = \left(\Sigma^{xw}_k\right)\T = \F_{k,:} \Phi(\vec{y})
\Sigma^{ww}_0
  - \F_{k,:} \left( \P + \K + \Q \right)
  \G\inv\Phi(\vec{y})\Sigma^{ww}_0
\end{equation*}
\begin{equation*}
  \Sigma^{ww}_k = \Sigma^{ww}_0 - \Sigma^{ww}_0\Phi(\vec{y})\T
  \G\inv\Phi(\vec{y})\Sigma^{ww}_0
\end{equation*}
\end{subequations}
for the covariance and
\begin{subequations}
\begin{equation*}
  m_k = \F_{k,0} m_0 + \F_{k,:}(\P + \Q + \K)\G\inv \vec{y}
\end{equation*}
\begin{equation*}
  \bar w_k = \bar w_0 + \Sigma^{ww}_0\Phi(\vec{y})\T
  \G\inv\vec{y}
\end{equation*}
\end{subequations}
for the mean.

\section{Gradients and Hessians of Dynamics Functions}
\label{sec:appendix-B}

\subsection{Neural Network Basis Functions}

The neural network dynamics function is
\begin{equation*}
  f(x) = \sum_{i=1}^F v_i \s(w_i(x-b_i)), \qq \s(a) = \frac{1}{1 + e^{-a}},
\end{equation*}
with the well-known derivatives of the logistic
\begin{equation*}
  \frac{\de}{\de a} \s(a) = \s(a)(1-\s(a)),
  \qq \frac{\de^2}{\de a^2} \s(a) =
    \s(a)\left(1-\s(a)\left(3-2\s(a)\right)\right).
\end{equation*}

The gradient of $f(x)$ can easily found to be
\begin{equation*}
  \nabla f(x) =
  \begin{bmatrix}
    \sum_{i=1}^F v_i w_i \s(w_i(x-b_i))(1-\s(w_i(x-b_i))) \\
    \s(w_1(x-b_1)) \\
    \vdots \\
    \s(w_F(x-b_F)) \\
    v_1 (x-b_1) \s(w_1(x-b_1)) (1 - \s(w_1(x-b_1))) \\
    \vdots \\
    v_F (x-b_F) \s(w_F(x-b_F)) (1 - \s(w_F(x-b_F)))
  \end{bmatrix}.
\end{equation*}

The Hessian, written in parts, using $a_i = w_i x + b_i$, is:
\begin{subequations}
  \begin{align*}
    \nabla^2_x f(x) &= \sum_{i=1}^F v_i w_i^2 \s(a_i)(1-\s(a_i)(3-2\s(a_i))) \\
    \nabla_x \nabla_{v_i} f(x) &= w_i\s(a_i)(1-\s(a_i)) \\
    \nabla_x \nabla_{w_i} f(x) &= v_i\s(a_i)(1-\s(a_i))
      + (x-b_i)w_i v_i \s(a_i)(1-\s(a_i))((1-2\s(a_i))\\
    \nabla_{v_i} \nabla_{v_i} f(x) &= 0 \\
    \nabla_{v_i} \nabla_{w_i} f(x) &= (x-b_i)\s(a_i)(1-\s(a_i)) \\
    \nabla_{w_i} \nabla_{w_i} f(x) &=
      v_i(x-b_i)^2\s(a_i)(1-\s(a_i)(3-2\s(a_i))).
  \end{align*}
\end{subequations}

\subsection{Fourier Basis Functions}

The Fourier approximation to the dynamics function has the form
\begin{equation*}
  f(x) = \sqrt{\frac{2}{F}}\sum_{i=1}^{\nicefrac{F}{2}} v_{2i-1}
    \sin(\w_{2i-1} x) + v_{2i} \cos(\w_{2i} x).
\end{equation*}

The gradient of $f(x)$ can easily verified to be
\begin{equation*}
  \nabla f(x) =
  \begin{bmatrix}
    \sqrt{\frac{2}{F}}\sum_{i=1}^{\nicefrac{F}{2}}
      v_{2i-1} \w_{2i-1} \cos(\w_{2i-1} x)
      - v_{2i} \w_{2i} \sin(\w_{2i} x) \\
    \sqrt{\frac{2}{F}} \sin(\w_{1} x)\\
    \sqrt{\frac{2}{F}} \cos(\w_{2} x)\\
    \vdots \\
    \sqrt{\frac{2}{F}} \sin(\w_{F-1} x)\\
    \sqrt{\frac{2}{F}} \cos(\w_{F} x)
  \end{bmatrix}.
\end{equation*}

The Hessian, written in parts,
using $c = \sqrt{\nicefrac{2}{F}}$ for normalization, is:
\begin{subequations}
  \begin{align*}
    \nabla^2_x f(x) &= c \sum_{i=1}^{\nicefrac{F}{2}}
      - v_{2i-1} \w_{2i-1}^2 \sin(\w_{2i-1} x)
      - v_{2i} \w_{2i}^2 \cos(\w_{2i} x) \\
    \nabla_x \nabla_{v_i} f(x) &= \begin{cases}
    c \omega_i \cos(\omega_i x) & \q i \q \text{odd}\\
    - c \omega_i \sin(\omega_i x) & \q i \q \text{even}
    \end{cases}\\
    \nabla_{v_i} \nabla_{v_i} f(x) &= 0.
  \end{align*}
\end{subequations}

\subsection{Radial Basis Functions}

With radial basis function features, the dynamics function is
\begin{equation*}
  f(x) = \sum_{i=1}^F v_i \exp\left(-\frac{(x-c_i)^2}{2\lambda^2}\right).
\end{equation*}

The gradient of $f(x)$ is
\begin{equation*}
  \nabla f(x) =
  \begin{bmatrix}
    \sum_{i=1}^F v_i \exp\left(-\frac{(x-c_i)^2}{2\lambda^2}\right)
      \frac{(c_i-x)}{2\lambda^2}\\
    \exp\left(-\frac{(x-c_1)^2}{2\lambda^2}\right) \\
    \vdots \\
    \exp\left(-\frac{(x-c_F)^2}{2\lambda^2}\right)
  \end{bmatrix}.
\end{equation*}

The Hessian, written in parts, is:
\begin{subequations}
  \begin{align*}
    \nabla^2_x f(x) &= \sum_{i=1}^F v_i
      \exp\left(-\frac{(x-c_i)^2}{2\lambda^2}\right)
      \left[
        \left(\frac{c_i - x}{2\lambda^2}\right)^2 - \frac{1}{\lambda^2}
      \right]
      \\
    \nabla_x \nabla_{v_i} f(x) &= \exp\left(-\frac{(x-c_i)^2}{2\lambda^2}\right)
        \frac{c_i - x}{2\lambda^2}\\
    \nabla_{v_i} \nabla_{v_i} f(x) &= 0 \\
  \end{align*}
\end{subequations}

\vskip 0.2in
\bibliography{bibfile}

\end{document}